%% file: main.tex
\documentclass[10pt,twocolumn,letterpaper]{article}

\usepackage[pagenumbers]{cvpr} %

\input{preamble}

\definecolor{cvprblue}{rgb}{0.21,0.49,0.74}
\usepackage[pagebackref,breaklinks,colorlinks,allcolors=cvprblue]{hyperref}

\title{Solv3R: 3D Looped Transformer as a Multi-view Solver}

\author{
Jeong-gi Kwak$^{1}$ \quad
Andrea Tagliasacchi$^{2,3,4}$ \quad
Sho Kagami$^{5}$ \quad
Yuki Ono$^{6}$ \quad
Kwang Moo Yi$^{1}$\\[0.25cm]
{\normalfont\normalsize
$^{1}$University of British Columbia \quad
$^{2}$Wayve Labs \quad
$^{3}$Simon Fraser University
}\\[0.05cm]
{\normalfont\normalsize
$^{4}$University of Toronto \quad
$^{5}$Sony Semiconductor Solutions Corporation \quad
$^{6}$Sony Corporation}
}

\makeatletter\patchcmd{\@maketitle}{\vspace*{12pt}}{}{}{}\makeatother
\begin{document}
\maketitle
\input{fig/intro_teaser}
\input{sec/0_abstract}

\input{sec/1_intro}

\input{sec/2_related}

\input{sec/3_method}
\input{sec/4_solver}

\input{sec/5_results}

\input{sec/6_conclusion}

\paragraph{Acknowledgments}
This work was supported in part by the Sony Faculty Innovation Award; the Natural Sciences and Engineering Research Council of Canada (NSERC) Discovery Grant and Alliance Advantage Grant; the Digital Research Alliance of Canada; and Advanced Research Computing at the University of British Columbia.
We thank Dmytro Mishkin for valuable feedback on the paper.

\clearpage
{
    \small
    \bibliographystyle{ieeenat_fullname}
    \bibliography{main}
}

\input{sec/X_suppl}

\end{document}

%% file: preamble.tex
\usepackage{cuted}
\usepackage{currfile}
\usepackage{caption}

\definecolor{turquoise}{cmyk}{0.65,0,0.1,0.3}
\definecolor{purple}{rgb}{0.65,0,0.65}
\definecolor{darkpurple}{rgb}{0.4,0,0.4}
\newcommand{\SupplementaryMaterial}{\textcolor{darkpurple}{supplementary material}}
\definecolor{dark_green}{rgb}{0, 0.5, 0}
\definecolor{orange}{rgb}{0.8, 0.6, 0.2}
\definecolor{red}{rgb}{0.8, 0.2, 0.2}
\definecolor{darkred}{rgb}{0.6, 0.1, 0.05}
\definecolor{blueish}{rgb}{0.0, 0.3, .6}
\definecolor{light_gray}{rgb}{0.7, 0.7, .7}
\definecolor{pink}{rgb}{0.9, 0, 0.6}
\definecolor{greyblue}{rgb}{0.25, 0.25, 1}
\definecolor{teal}{rgb}{0.0, 0.4, 0.4}

\definecolor{dark_green}{rgb}{0, 0.4, 0}


\usepackage{fvextra}
\usepackage[most]{tcolorbox}
\usepackage{booktabs}
\usepackage{multirow}
\usepackage{url}

\usepackage{colortbl}
\definecolor{rkfirst}{RGB}{248,212,212}
\definecolor{rksecond}{RGB}{252,227,206}
\definecolor{rkthird}{RGB}{252,244,209}
\colorlet{color1}{rkfirst}
\colorlet{color2}{rksecond}
\colorlet{color3}{rkthird}
\newcommand{\best}[1]{\cellcolor{rkfirst}#1}
\newcommand{\second}[1]{\cellcolor{rksecond}#1}
\newcommand{\third}[1]{\cellcolor{rkthird}#1}

\renewcommand{\paragraph}[1]{\vspace{.2em}\noindent\textbf{#1}}

\newcommand{\dejaview}{D\'ej\`aView\xspace}

\newlist{properties}{enumerate}{1}
\setlist[properties]{label=\textbf{(P\arabic*)},ref=P\arabic*,wide=0pt,labelsep=0.3em,noitemsep,topsep=1pt}
\AtEndPreamble{\crefformat{propertiesi}{#2#1#3}\Crefformat{propertiesi}{#2#1#3}\crefrangeformat{propertiesi}{#3#1#4--#5#2#6}\crefmultiformat{propertiesi}{#2#1#3}{ and #2#1#3}{, #2#1#3}{, and #2#1#3}}

\usepackage{soul}
\setuldepth{foobar}

\makeatletter
\patchcmd{\NAT@citexnum}
  {{\reset@font\bfseries?}}
  {{\reset@font\bfseries\textcolor{red}{\detokenize\expandafter{\@citeb}}}\ignorespaces}
  {}
  {\PackageWarning{preamble}{Could not patch natbib's undefined-citation output}}
\makeatother

%% file: fig/intro_teaser.tex
\begin{strip}
\centering
\includegraphics[width=0.96\linewidth]{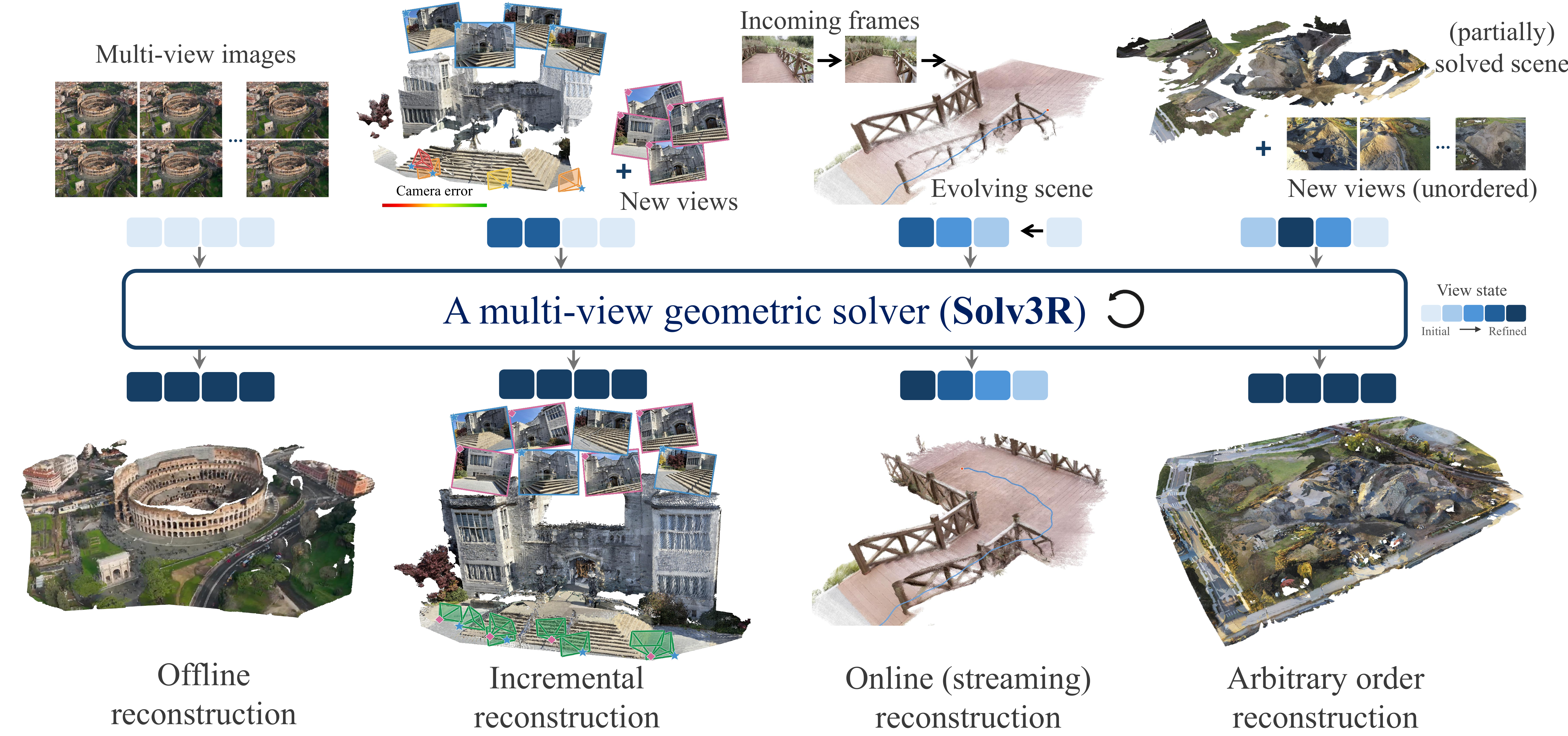}
\captionsetup{hypcap=false}
\captionof{figure}{\textbf{Teaser.}
We present Solv3R, a looped transformer that turns multi-view reconstruction into a continued solving process.
The same learned solver continues from its current state, jointly refining previously reconstructed and newly added views.
This allows reconstruction to evolve as views are introduced at any stage or in any order, without restarting.
}
\captionsetup{hypcap=true}
\label{fig:\currfilebase}
\end{strip}

%% file: sec/0_abstract.tex
\begin{abstract}
Classical structure-from-motion reconstruction proceeds iteratively, refining camera poses and scene geometry from a current estimate.
This allows maps to easily expand without recomputing solved views, or geometry estimates to improve as more observations are provided.
Most geometric foundation models, however, predict entire geometry in a single forward pass or update a recurrent state as frames arrive sequentially, making this expansion and refinement non-trivial.
We introduce Solv3R, a looped transformer that learns to continue solving multi-view geometry from its current state.
Importantly, its update operator is independent of the iteration count, allowing previously refined and newly introduced views to be updated jointly or selectively, regardless of how ``solved'' they are.
We train the operator to preserve accurate reconstructions under continued iteration and correct inaccurate states, enabling refinement of fixed image collections and integration of new observations.
Representations of solved views can also be cached once and reused as geometric context across iterations, enabling compact and efficient key-value caching.
We analyze the solver dynamics and compare it with the state-of-the-art, showing that a single learned solver supports offline and incremental reconstruction, relocalization, and streaming without task-specific training or learned modules.
\iftoggle{cvprfinal}{%
The code and model are available at \url{https://github.com/ubc-vision/solv3r}.}{The code and model are available in the supplementary material, and will be publicly released.}
\end{abstract}

%% file: sec/1_intro.tex
\section{Introduction}
\label{sec:intro}
\input{fig/intro_update}

Recent geometric foundation models (GFMs)~\cite{dust3r,mast3r,fast3r,pi3,vggt,depthanything3,vggtomega} have driven a shift in multi-view 3D reconstruction toward predicting scene geometry directly from multi-view images.
Unlike classical Structure-from-Motion~(SfM), which relies on correspondence matching and \textit{iterative} geometric optimization~\cite{schonberger2016sfm}, these models estimate geometry in a \textit{feed-forward} manner.
They typically employ successive transformer blocks to progressively build geometry-aware representations and predict camera poses, depths, and point maps.

But large-scale reconstruction often extends beyond a single forward pass.
The image collection may exceed the input capacity of a transformer, or new views may arrive after the initial reconstruction, such as subsequent video frames or additional images and videos of the same scene.
Incorporating these views with a feed-forward GFM typically requires processing the combined image set again~\cite{vggt,pi3}.
Streaming models~\cite{spann3r,cut3r,streamvggt,stream3r,ttt3r} avoid this by updating a state frame by frame, but they are limited to views that arrive \textit{sequentially}.
In contrast, classic SfM incorporates new observations into an existing reconstruction and uses \textit{iterative} optimization to refine its estimates~\cite{schonberger2016sfm}.
Hence, a learned reconstruction model should likewise be able to \textit{continue} solving from its current state, refining existing estimates and integrating new views as they arrive.
This requires a reconstruction ``state'' that remains usable for further geometric computation beyond the initial forward pass.

But conventional feed-forward GFMs like VGGT~\cite{vggt} process image representations through successive transformer layers, each applying a \textit{different} learned transformation (\cref{fig:intro_update}).
Their intermediate states (i.e.~tokens) are consequently tied to particular \textit{stages} of computation (i.e.~layer depth).
\dejaview~\cite{burzio2026dejaview} moves toward iterative reconstruction by repeatedly applying a shared transformer block, but conditions its updates on a planned schedule.
Neither is designed to jointly update views with different iteration histories.
In contrast, a classical \textit{stationary} iterative solver such as Jacobi~\cite{barrett1994templates} applies the same update at every iteration, and \textit{does not} need to know which iteration it is operating on.
It can thus continue from any state, including one whose variables have received different numbers of iterations.
A key observation is that sharing weights alone \textit{does not} give \dejaview this ability.

We thus posit that a GFM \textit{should be a learning-based multi-view solver}.
Given more computation, a solver should continue refining a reconstruction until convergence, then preserve the solution under further iterations.
A solver of this type should handle offline reconstruction, adding views, and streaming with one unifying architecture (see \cref{fig:intro_teaser}).

To this end, we introduce Solv3R, an architecture that reformulates multi-view reconstruction as a reusable learning-based geometric solver.
Solv3R employs a stationary (\ie iteration-independent) learned geometric update operator that acts on the current multi-view state.
We train this operator on diverse solver states so that accurate reconstructions remain stable under further iterations, while inaccurate states and newly added views can be refined through continued solving (\cref{fig:intro_incremental}).

With these solver properties, Solv3R turns multi-view reconstruction from a fixed-set prediction into a \textit{continued solving process}.
Our solver achieves competitive accuracy in conventional multi-view reconstruction, while allowing existing estimates to be jointly revised as new observations arrive.
Solv3R can also reconstruct an entire scene by iteratively updating only small subsets of views, rather than updating all views jointly.
This reuse extends beyond reconstruction states to attention context: a fixed reconstructed map can provide a single KV cache reused across solver calls.
Together, these capabilities support incremental reconstruction, map-based relocalization, and streaming reconstruction without task-specific learned modules, achieving competitive streaming performance even without streaming-specific training.

Through extensive analyses of solver dynamics and experiments across reconstruction scenarios, we demonstrate how reusable geometric computation enables flexible and efficient multi-view reconstruction.

\input{fig/intro_incremental}

%% file: fig/intro_update.tex
\begin{figure*}[t]
\centering
\includegraphics[width=\linewidth]{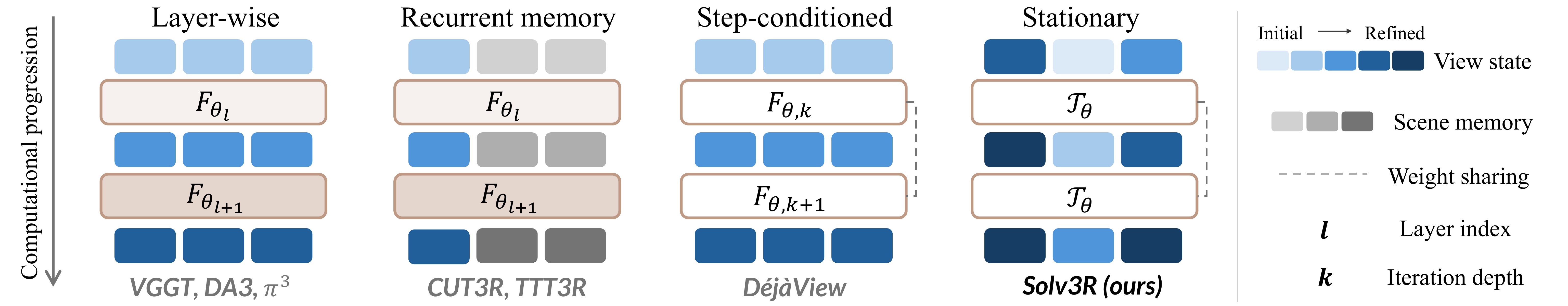}
\caption{
Comparison of geometric update mechanisms.
Conventional GFMs~\cite{vggt,depthanything3,pi3} use successive layer-specific transformations, tying their intermediate representations to layer depth.
Recurrent-memory models~\cite{cut3r,ttt3r} process incoming frames while updating a persistent scene memory across observations.
Although \dejaview~\cite{burzio2026dejaview} shares transformer weights, its updates remain conditioned on a planned schedule.
Solv3R applies the same stationary update operator regardless of the iteration count, allowing previously refined and newly introduced views to be jointly updated.
}
\label{fig:\currfilebase}
\end{figure*}

%% file: fig/intro_incremental.tex
\begin{figure}[t]
\centering
\includegraphics[width=\linewidth]{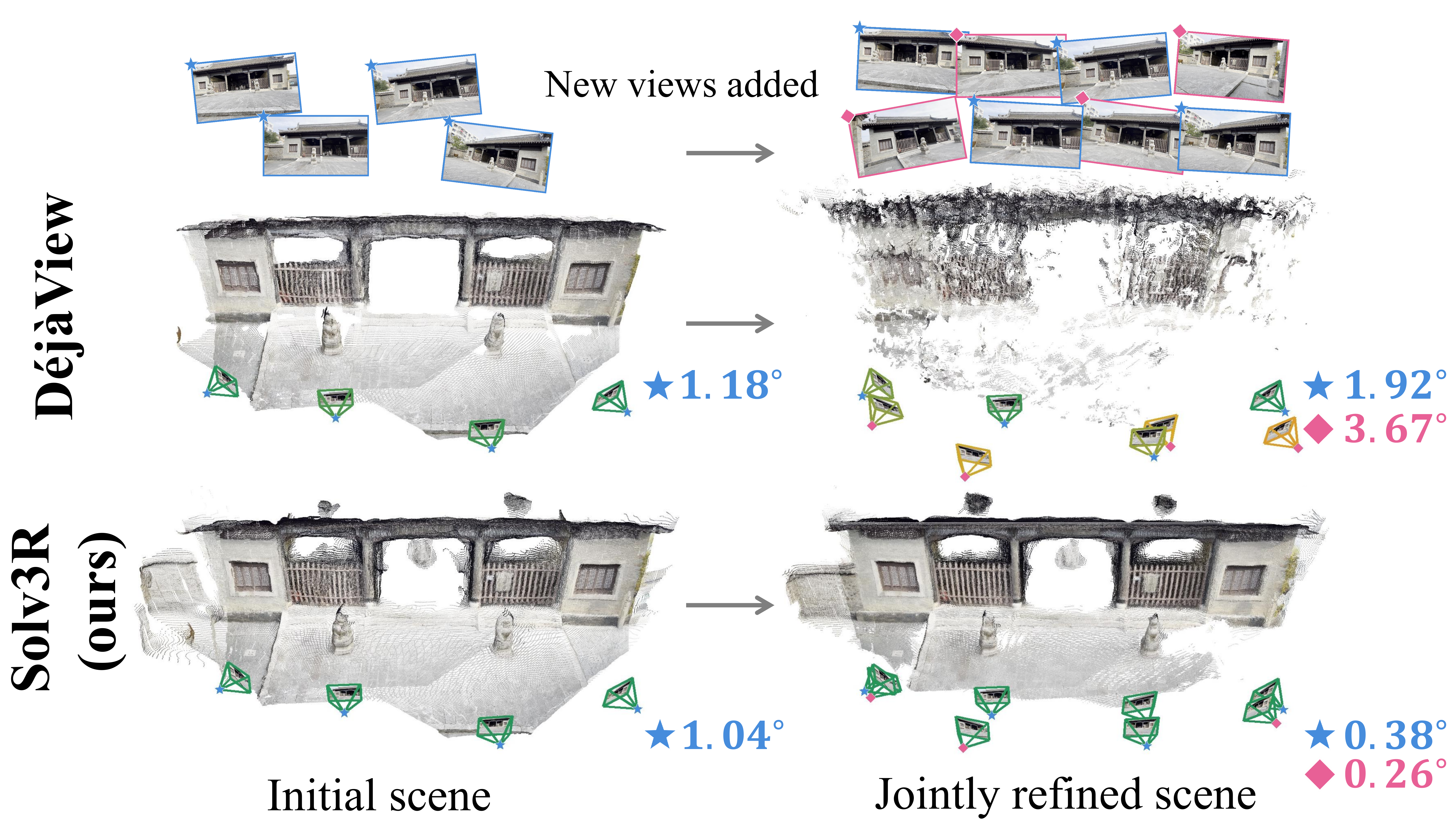}
\caption{
Continuing geometric solving with new observations.
When new views are introduced, Solv3R continues solving from its existing state, jointly refining both previously reconstructed and newly added views.
\dejaview struggles to maintain consistent reconstruction.
Numbers indicate camera rotation errors in degrees.
}
\label{fig:\currfilebase}
\end{figure}

%% file: sec/2_related.tex
\section{Related work}

Geometric foundation models (GFMs) have advanced feed-forward 3D reconstruction, progressing from pairwise pointmap regression and matching~\cite{dust3r,mast3r} to joint prediction of camera poses and dense geometry from multiple views~\cite{mvdust3r,fast3r,must3r,light3rsfm,vggt,pi3,depthanything3}.
Recent methods further improve multi-view geometric prediction and scalability~\cite{vggtomega,amb3r,dens3r,flare,more,dage,carve,hdvggt,nova3r,sage}.
Related advances extend geometric prediction to monocular inputs~\cite{depthanythingv2,moge,moge2,metric3dv2}, incorporate additional geometric priors~\cite{pow3r,mapanything,omnivggt,worldmirror,matrix3d}, or produce renderable Gaussian representations~\cite{mvsplat,noposplat,depthsplat,yonosplat}.

\paragraph{Learned iterative refinement and solvers.}
Learned iterative refinement has been widely applied to optical flow, stereo, multi-view depth, visual SLAM, and bundle adjustment~\cite{raft,raftstereo,itermvs,droidslam,deqflow,bat}.
Related work on equilibrium and recurrent models investigates fixed-point behavior, stability, and extended computation~\cite{deq,pathindependent,overthinking,streamdeq}.
Inspired by recent advances in looped transformers for language models~\cite{universaltransformer,cotformer,relaxedrecursive,huginn,mixtureofrecursions,ouro,loopformer,modr}, \dejaview~\cite{burzio2026dejaview} introduces shared recurrent computation for multi-view reconstruction.
We extend this direction to continued geometric solving across fixed and evolving view sets, including views at different refinement stages.

\paragraph{Continuous and streaming 3D reconstruction.}
Continuous reconstruction methods maintain and update geometry as new observations arrive.
Some approaches employ persistent scene representations or test-time adaptation~\cite{spann3r,cut3r,point3r,ttt3r,loger,wint3r}, while others preserve geometric context through causal attention and caching~\cite{streamvggt,stream3r,infinitevggt,lingbotmap,slamformer}.
Window-based methods further integrate local reconstructions through registration and optimization~\cite{slam3r,mast3rslam,vggtslam,vggtlong,talo,suremap}.
Rather than introducing a separate learned reconstruction process for streaming, Solv3R applies the same geometric solver to both offline and online reconstruction.

\input{fig/method}

%% file: fig/method.tex
\begin{figure*}[t]
\centering
\includegraphics[width=0.95\linewidth]{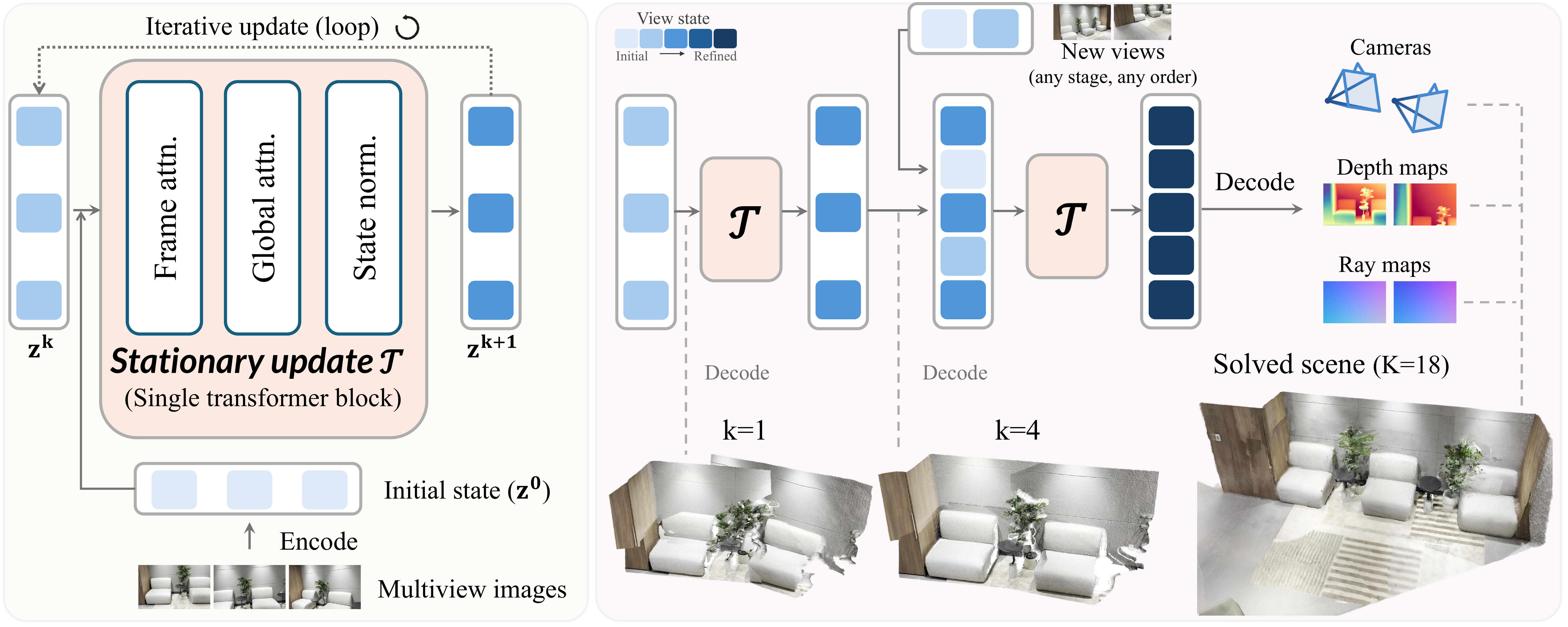}
\caption{
\textbf{Overview.}
Left: A stationary transformer block repeatedly updates the multi-view state through frame-wise and global attention, initial-state injection, and state normalization.
Right: The same operator continues refining the reconstruction across solver calls, allowing new views to be introduced at any stage and jointly updated with existing views.
Camera poses, depth maps, and ray maps can be decoded from the current state after any solver call.
}
\label{fig:\currfilebase}
\end{figure*}

%% file: sec/3_method.tex
\section{Method}
\label{sec:method}

Our key idea is to make the shared block of a looped transformer act as the ``iteration'' of a classical solver, so that it continues solving from the current state, regardless of how many iterations each view has received.
We review looped transformers for 3D reconstruction in \cref{sec:preliminaries}, and use a simple classical solver in \cref{sec:solver_view} to define the properties that we want from such a solver.
We then replace the classical iteration with a learned operator in \cref{sec:geometric_solver}, and describe how we train it to exhibit these properties in \cref{sec:solver_training}.

\subsection{Looped transformers for 3D reconstruction}
\label{sec:preliminaries}
\dejaview~\cite{burzio2026dejaview} formulates multi-view reconstruction as iterative refinement of a latent state.
Given input images $\mathcal{I}{=}\{I_i\}_{i=1}^{N}$, an image encoder initializes a multi-view state $\mathbf{z}^0$, which is repeatedly updated by a shared transformer block with frame-wise and global attention.
The refined state is decoded into depth and ray maps, from which scene geometry and camera poses are recovered.
Unlike conventional feed-forward transformers with independently parameterized layers, \dejaview \textit{shares} the transformer block across iterations, but conditions each update on the step:
\begin{equation}
    \mathbf{z}^{k+1} = F_\theta(\mathbf{z}^k; \tau_k),
    \label{eq:dvlt_update}
\end{equation}
where $\tau_k$ is the conditioning on step $k$ of a planned schedule, which lets the shared block apply a \textit{different} transformation at each iteration.
Because of this structure, \dejaview loses reconstruction accuracy when iterated beyond its trained range, as we demonstrate in \Cref{sec:solver_properties}.
More fundamentally, each state is tied to the iteration index at which it was computed.
One application of $F_\theta$ thus cannot jointly update views that have received different numbers of iterations, because no single $\tau_k$ matches all of them.
Thus, sharing weights alone \textit{does not} give the block an update that is independent of the iteration index.

\subsection{An iterative solver as a running example}
\label{sec:solver_view}
To make the properties we want concrete, consider a classical constraint satisfaction problem, solving a linear system $A\mathbf{x}{=}\mathbf{b}$ with the Jacobi method~\cite{barrett1994templates}.
The rows of matrix $A$ and corresponding observations $\mathbf{b}$ define the constraints, and the state $\mathbf{x}$ holds the current estimate of the unknowns.
Starting from an initial state $\mathbf{x}^0$, each iteration applies the same update:
\begin{align}
    \mathbf{x}^{k+1} &= \Phi(\mathbf{x}^{k};A,\mathbf{b}),
    \label{eq:jacobi}
\end{align}
We call one iteration, that is, one application of $\Phi$, a solver call.
The update $\Phi$ does not depend on the iteration index $k$, which makes the Jacobi method \textit{stationary}~\cite{barrett1994templates}.
When the method converges, three \textit{properties} follow:
\begin{properties}
    \item \label{prop:stability}\textit{Stability.}
    The accurate solution $\mathbf{x}^*$, which satisfies $A\mathbf{x}^*{=}\mathbf{b}$, is a fixed point with $\Phi(\mathbf{x}^*;A,\mathbf{b}){=}\mathbf{x}^*$, so further solver calls leave it unchanged.
    \item \label{prop:correction}\textit{Correction.}
    Consider a perturbed state, that is, a solution moved off by, for example, added noise.
    From such a state, the same solver calls converge back to $\mathbf{x}^*$.
    \item \label{prop:continuation}\textit{Continuation.}
    When new unknowns and constraints are added, the solver can start the old unknowns from their solved values, and the new ones from an initial guess.
    It then updates all of them together toward the solution of the enlarged system, often in fewer solver calls than a restart from $\mathbf{x}^0$.
    The state holds values that have received different numbers of solver calls, and $\Phi$ still updates them in one solver call.
    It can do so because it does not depend on how a state was reached.
\end{properties}

Multi-view reconstruction has the same structure.
The input images define the constraints, the per-view states hold the current estimate, and the cameras and depths decoded from the state are the current estimate of the solution.
A solved state is one whose decoded cameras and depths are accurate, and which further solver calls change little.
A learned multi-view solver should thus leave a solved reconstruction unchanged under further solver calls {(\cref{prop:stability})}, correct a perturbed reconstruction {(\cref{prop:correction})}, and continue from a solved reconstruction when views are added {(\cref{prop:continuation})}.
Meanwhile, the update of \dejaview in \cref{eq:dvlt_update} depends on $\tau_k$, so it \textit{does not} have these properties by construction.

\input{fig/stability}

\subsection{A stationary geometric solver}
\label{sec:geometric_solver}
We replace the update $\Phi$ of \cref{eq:jacobi} with a learned operator $\mathcal{T}_\theta$, and the constraints $(A,\mathbf{b})$ with the encoded input images $\mathbf{z}^0{=}E(\mathcal{I})$, as illustrated in~\Cref{fig:method}:
\begin{equation}
\mathbf{z}^{k+1} = \mathcal{T}_\theta(\mathbf{z}^{k};\mathbf{z}^{0}),
\label{eq:stationary_solver}
  \end{equation}
where the state also starts from the \textit{initial} encoding $\mathbf{z}^0$.
Like $\Phi$, the operator is \textit{stationary} (\ie iteration-independent).
It needs neither the iteration index nor the number of solver calls that each view has received.
The state may thus combine views with different numbers of solver calls, $\mathbf{z}^k{=}(\mathbf{z}_1^{[h_1]},\ldots,\mathbf{z}_N^{[h_N]})$, where view $i$ has received $h_i$ calls.
We inject $\mathbf{z}^0$ at every solver call for the same reason that $\Phi$ reads $A$ and $\mathbf{b}$ at every call.
The state tracks the current estimate, while $\mathbf{z}^0$ carries the constraints from the images.
Attention infers how the views constrain each other.
Without $\mathbf{z}^0$, each call would see only the current state, and errors in the state could accumulate over calls.
We instantiate $\mathcal{T}_\theta$ as a \textit{single} transformer block that is completely \textit{shared} across solver calls, with frame-wise and global attention.

In the looped block, the step conditioning $\tau_k$ enters only through gates that scale the channels of each sub-block.
To make the update independent of the iteration index, we replace them with learned \textit{constant} gates, which apply the same scaling at every solver call.
Since no schedule bounds the number of solver calls, we keep the magnitude of the state from drifting by rescaling the updated state to a fixed root-mean-square (RMS) norm.
We use a single scale factor for the whole state, which preserves the relative magnitudes across views.
Stationarity alone does not make $\mathcal{T}_\theta$ converge.
We thus train it to exhibit \cref{prop:stability,prop:correction,prop:continuation}, as detailed in \cref{sec:solver_training}.

\subsection{Learning solver dynamics}
\label{sec:solver_training}
We train $\mathcal{T}_\theta$ to exhibit \cref{prop:stability,prop:correction,prop:continuation}.
When the input images do not change, a solved state $\mathbf{z}^*$ should change little under further solver calls, $\mathcal{T}_\theta(\mathbf{z}^*;\mathbf{z}^0){\approx}\mathbf{z}^*$.
From a perturbed state, the same operator should bring the reconstruction back to the solution.
We thus train it on states that have received different numbers of solver calls, as well as on perturbed states.

\paragraph{Training states.}
To keep solved states unchanged under further solver calls (\cref{prop:stability}), we train the operator on heterogeneous states.

In the main training stage, starting from $\mathbf{z}^0$, we apply a random number of solver calls without gradients, followed by several differentiable calls with supervision at the final call.
This exposes the operator to states at different refinement stages, encouraging stable updates under continued computation.
To train the operator to correct perturbed states (\cref{prop:correction}), we occasionally perturb the state before the differentiable calls.
We either add Gaussian noise to the whole state or reset a subset of views to $\mathbf{z}^0$.
Gaussian noise mimics deviations from a solution, while partial resets simulate newly introduced views alongside refined ones~(\cref{prop:continuation}).
Similar in spirit to denoising-based learning~\cite{vincent2011connection,song2019generative} and Diffusion Forcing~\cite{diffusionforcing}, these perturbations expose the operator to a broader range of states, encouraging corrective refinement beyond its usual trajectories.

\paragraph{Geometric supervision.}
To encourage accurate geometric predictions throughout continued solving, we apply supervision after varying numbers of solver calls.
At the final call of each sampled rollout (24--48 calls), we decode camera poses, ray maps, and depths and apply ground-truth supervision $\mathcal{L}_{\mathrm{geo}}$ where available.
To preserve the priors of the pretrained \dejaview, we also distill its predictions after 16 solver calls using $\mathcal{L}_{\mathrm{dist}}$.

The overall training objective is
\begin{equation}
    \mathcal{L}
    =
    \mathcal{L}_{\mathrm{geo}}
    +
    \lambda_{\mathrm{dist}}
    \mathcal{L}_{\mathrm{dist}}.
    \label{eq:solver_loss}
\end{equation}
We provide further details on the loss terms and training procedure in the \SupplementaryMaterial{}.

%% file: fig/stability.tex
\begin{figure*}[t]
\centering
\includegraphics[width=\linewidth]{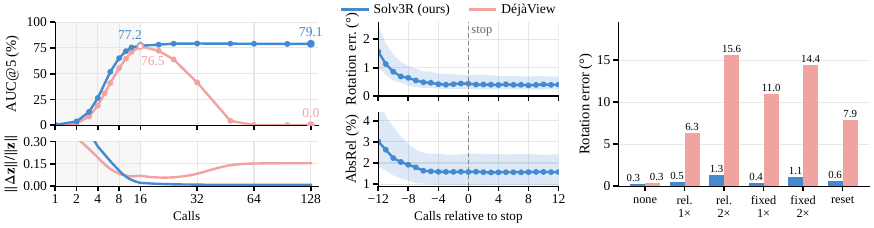}
\caption{\textbf{Stable and corrective solver dynamics.}
\ul{Left}: pose accuracy and relative state update norm over solver calls.
\dejaview is trained with up to 16 steps (shaded).
\ul{Middle}: camera rotation error and depth AbsRel aligned to the solver call at which each window meets the stopping criterion of \cref{eq:relative_update}, showing median and interquartile range.
\ul{Right}: rotation error at call 16 after perturbations at call 8 using relative (rel.) or fixed-scale (fixed) Gaussian noise, or state resets (reset).
These results show that Solv3R preserves accurate reconstructions under continued computation, and that its state stabilization coincides with stabilization in geometric accuracy.
}
\label{fig:\currfilebase}
\end{figure*}

%% file: sec/4_solver.tex
\section{Solver analysis}
\label{sec:solver_properties}
We first evaluate whether Solv3R exhibits the three solver properties defined in \Cref{sec:solver_view}, comparing it with \dejaview~\cite{burzio2026dejaview}.

\paragraph{Stability (\cref{prop:stability}) and correction (\cref{prop:correction}) -- \Cref{fig:stability}.}
We examine whether our solver preserves accurate reconstructions under continued solver calls and recovers from perturbed states.
To measure solver stabilization, we define the relative state update between consecutive calls as
\begin{equation}
r^k =
\frac{\|\mathbf{z}^{k+1}-\mathbf{z}^{k}\|_2}
{\|\mathbf{z}^{k}\|_2}.
\label{eq:relative_update}
\end{equation}

As shown in \Cref{fig:stability}~(left), our solver maintains reconstruction accuracy as its relative state updates diminish, whereas \dejaview loses accuracy under continued computation.
We use this measure for adaptive inference, stopping refinement when $r^k < \gamma$.
We empirically select $\gamma{=}0.02$ to balance reconstruction accuracy and computational cost.
Importantly, a small state update does not necessarily imply geometric accuracy.
We therefore examine whether state stabilization coincides with stabilization of geometric error.
As shown in \Cref{fig:stability}~(middle), rotation and depth errors stabilize as the state updates diminish and remain low under further refinement.
Among the 144 windows meeting the stopping criterion, 143 have rotation errors below $5^\circ$.
The median error changes only from $0.44^\circ$ at stopping to $0.43^\circ$ after 64 calls.
Similar behavior is observed across additional datasets (see \SupplementaryMaterial{}).
These results demonstrate that state stabilization corresponds to stabilization in geometric accuracy.

\input{fig/bridge}
To test corrective behavior, we perturb the state after eight solver calls by adding Gaussian noise or resetting views to their initial states, then continue refinement for eight more calls.
As shown in \Cref{fig:stability}~(right), our solver recovers accurate camera poses from every perturbation, whereas \dejaview does not. The solver also reaches nearly identical camera estimates from different initial states after extended refinement (see \SupplementaryMaterial{}).

\paragraph{Continuation with new constraints (\cref{prop:correction,prop:continuation}) -- \Cref{fig:bridge}.}
We examine whether new geometric constraints can correct an existing inaccurate reconstruction through continued solving.
We construct challenging reconstruction windows with two groups of views sharing very limited overlap.
As shown in \Cref{fig:bridge}, both Solv3R and \dejaview achieve low rotation errors when re-solving with bridging views from scratch.
However, only our method corrects the original estimates through continued solving, reaching nearly the same accuracy as re-solving from scratch.
With 16 bridging views, all 17 windows recover to below $5^\circ$, whereas \dejaview fails to recover any.
These results demonstrate that the solver can incorporate new geometric constraints to revise an existing reconstruction without restarting the solve.

\input{fig/warm_start}

\paragraph{Warm-start continuation (\cref{prop:continuation}) -- \Cref{fig:warm_start}.}
When a new frame arrives, a reusable solver should retain the states of previously reconstructed views rather than restarting the entire reconstruction.
In a sliding-window setting, we preserve the current states of overlapping views while initializing newly introduced views from their image encodings.
As shown in \Cref{fig:warm_start}, Solv3R maintains stable reconstruction accuracy over repeated warm starts, whereas \dejaview diverges.
Additionally, our method reaches accurate camera estimates in fewer solver calls than restarting from scratch, demonstrating that previously computed states can be effectively reused.

\input{fig/mixed_progress}

\paragraph{Mixed-progress solving (\cref{prop:continuation}) -- \Cref{fig:mixed_progress}.}
A reusable multi-view solver should jointly update views at different refinement stages without requiring synchronized progress.
We evaluate this capability through experiments in \Cref{fig:mixed_progress}.
In \Cref{fig:mixed_progress}~(left), we initialize each view from randomly selected reconstruction stages (0--16 calls).
Despite having input in varying stages, our solver produces nearly identical camera rotations across independent trials, with a mean difference of only $0.08^\circ$.
In contrast, \dejaview remains sensitive to initialization as simply using a looped transformer does not result in solver-like properties.

\paragraph{Enhanced throughput via pipelining (\cref{prop:continuation}) -- \Cref{fig:mixed_progress}.}
This mixed-progress solving capacity allows us to enhance our throughput.
In \Cref{fig:mixed_progress}~(right) we further test a \emph{pipelined} setup, where the sliding window advances after every solver call rather than waiting for convergence.
In other words, each view in the window has an increasing number of calls, with the most recent view having a single call, and the oldest view in the window having the most calls.
Solv3R progressively refines camera estimates properly despite this non-conventional setup.
It is important to point out that this means that, once the first, \eg, 16 calls are made, the delay between estimates for each frame is now a \emph{single call}, because frames are already being processed while previous frames are being estimated.
This effectively enhances the throughput of our solver by the window size (note that the latency of estimates remains the same).
For example, compared to \dejaview, which must re-solve the window for every new frame, we share the same latency when we use 16 calls, but our solver processes 16$\times$ more frames per second (10$\times$ end to end).

\paragraph{Reusable geometric context~(\cref{prop:stability}).}
The properties of our method also lead to efficient \textit{caching}.
For efficient inference, key--value (KV) caching allows previously reconstructed views to serve as fixed attention context without repeated computation.
In conventional stacked GFMs, these representations are layer-specific and must be cached separately for each transformer layer.
Although \dejaview shares transformer weights across iterations, its step-dependent updates \textit{limit} KV cache reuse across refinement stages.
By contrast, our solver reuses a single KV cache computed from fixed views across all solver calls, regardless of the refinement stage of the query.
This extends reuse beyond model parameters to the results of geometric computation, reducing KV cache storage by $16\times$ compared with stage-specific caching in \dejaview.
We demonstrate this capability applied to map-based relocalization  in~\Cref{sec:relocalization}.

\input{fig/blockwise}

\input{tab/offline_reconstruction}

\paragraph{Block-wise geometric solving (\cref{prop:stability,prop:continuation}) -- \Cref{fig:blockwise}.}
Mixed-progress solving and reusable geometric context allow Solv3R to refine a global reconstruction without updating all views jointly.
At each solver call, we update a randomly selected block of views while using the remaining views as fixed geometric context, as in block-coordinate methods~\cite{wright2015coordinate}.
We evaluate this on 20 scenes with 128 views each, updating only eight views per call.
As shown in \Cref{fig:blockwise}, our model approaches the same camera pose and geometry accuracy as joint solving under continued block-wise updates.
In contrast, the accuracy of \dejaview degrades significantly under random block-wise updates, even with the same number of updates per view.
By updating only a small block at each call, this approach reduces peak GPU memory compared with joint solving, requiring $10\times$ less than VGGT for 128 views (1.38 vs.\ 13.98 GiB).

\input{fig/incremental_join}
\input{fig/incremental_efficiency}

%% file: fig/bridge.tex
\begin{figure}[t]
\centering
\includegraphics[width=0.95\linewidth]{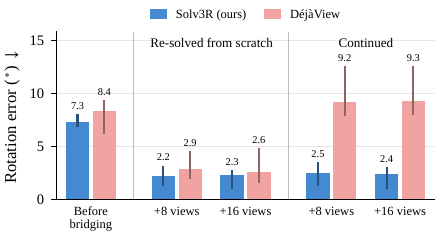}
\caption{
\textbf{Correction with new observations.}
Camera rotation error on the original views across 17 weak-overlap windows (median and interquartile range).
\emph{Before bridging} shows the reconstruction before adding bridging views.
\emph{Re-solved from scratch} and \emph{Continued} use the same 8 or 16 bridging views, either restarting the reconstruction or continuing from the existing states.
Both models achieve low errors when re-solving from scratch, but only ours can correct the original view estimates through continued solving.
}
\label{fig:\currfilebase}
\end{figure}

%% file: fig/warm_start.tex
\begin{figure}[t]
\centering
\includegraphics[width=\linewidth]{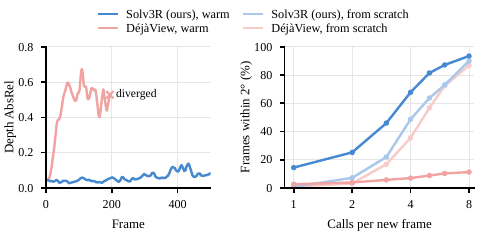}
\caption{\textbf{Warm-start convergence.}
\ul{Left}: depth AbsRel along a KITTI~\cite{kitti} sequence, warm-starting each new frame from the previous solution with four solver calls.
\ul{Right}: percentage of new frames with rotation error below $2^\circ$ versus solver calls in a 24-view window, comparing warm-starting with restarting from scratch.
Ours remains stable under continued warm-starting and achieves accurate camera estimates with fewer calls, whereas \dejaview diverges.}
\label{fig:\currfilebase}
\end{figure}

%% file: fig/mixed_progress.tex
\begin{figure}[t]
\centering
\includegraphics[width=\linewidth]{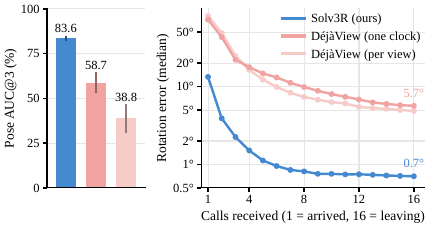}
\caption{\textbf{Mixed-progress solving.}
\ul{Left}: pose accuracy when each view is initialized from a randomly sampled reconstruction stage.
\ul{Right}: rotation error as each view progresses through a pipelined reconstruction window, where views enter and leave after each solver call.
For \dejaview, \textit{one clock} uses a shared iteration index, while \textit{per view} uses the iteration count of each view.
Our solver maintains accurate reconstruction across views with varying reconstruction stages, whereas \dejaview is sensitive to their computational progress.}
\label{fig:\currfilebase}
\end{figure}

%% file: fig/blockwise.tex
\begin{figure}[t]
\centering
\includegraphics[width=\linewidth]{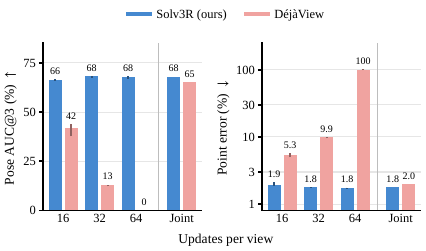}
\caption{
\textbf{Block-wise geometric solving.}
We evaluate 128-view reconstruction by updating eight randomly selected views per solver call.
\ul{Left}: Pose AUC@3 under block-wise updates, compared with joint solving.
\ul{Right}: Point error under the same settings.
The remaining views provide cached geometric context.
Ours approaches joint-solving accuracy in both camera poses and global geometry, whereas \dejaview degrades as refinement proceeds without synchronized updates.
}
\label{fig:\currfilebase}
\end{figure}

%% file: tab/offline_reconstruction.tex
\newcommand{\tabvenue}[1]{%
  \hspace{2pt}{\color{gray}\fontsize{5.7}{6.8}\selectfont (#1)}%
}

\newcommand{\tabparams}[2]{%
  #1\,\makebox[2.9em][l]{\ifx\relax#2\relax\else{\color{gray}\fontsize{5.7}{6.8}\selectfont(#2$\times$)}\fi}%
}

\begin{table*}[h]
\begin{center}
\resizebox{\textwidth}{!}{%
\setlength{\tabcolsep}{2.6pt}%
\setlength{\fboxsep}{1pt}%
\renewcommand{\arraystretch}{1.23}%
\fontsize{7.0}{8.4}\selectfont
\begin{tabular}{@{}lrr*{10}{r}*{10}{r}}
\toprule
& \multicolumn{2}{c}{\textbf{Efficiency}} & \multicolumn{10}{c}{\textbf{Camera Pose}} & \multicolumn{10}{c}{\textbf{Geometry}} \\
\cmidrule(lr){2-3}\cmidrule(lr){4-13}\cmidrule(l){14-23}
& & & \multicolumn{2}{c}{DTU} & \multicolumn{2}{c}{ETH3D} & \multicolumn{2}{c}{7-Scenes} & \multicolumn{2}{c}{ScanNet++} & \multicolumn{2}{c}{nuScenes} & \multicolumn{2}{c}{DTU} & \multicolumn{2}{c}{ETH3D} & \multicolumn{2}{c}{7-Scenes} & \multicolumn{2}{c}{ScanNet++} & \multicolumn{2}{c}{nuScenes} \\
\cmidrule(lr){4-5}\cmidrule(lr){6-7}\cmidrule(lr){8-9}\cmidrule(lr){10-11}\cmidrule(lr){12-13}\cmidrule(lr){14-15}\cmidrule(lr){16-17}\cmidrule(lr){18-19}\cmidrule(lr){20-21}\cmidrule(l){22-23}
\textbf{Method} & \multicolumn{1}{c}{\textbf{Params}} & \textbf{TFLOPs} & \textbf{@3} & \textbf{@30} & \textbf{@3} & \textbf{@30} & \textbf{@3} & \textbf{@30} & \textbf{@3} & \textbf{@30} & \textbf{@3} & \textbf{@30} & \textbf{L2} & \textbf{IR} & \textbf{L2} & \textbf{IR} & \textbf{L2} & \textbf{IR} & \textbf{L2} & \textbf{IR} & \textbf{L2} & \textbf{IR} \\
\midrule

MASt3R~\cite{mast3r}\tabvenue{ECCV'24} &
\tabparams{689M}{5.9} & 500.0 &
21.6 & 81.7 & 35.2 & 57.6 & 9.3 & 70.6 & 15.5 & 43.9 & 9.4 & 62.7 &
.011 & 94.9 & .340 & 29.1 & .076 & 41.7 & .251 & 14.5 & .360 & 11.4 \\

MASt3R-SfM~\cite{mast3rsfm}\tabvenue{3DV'25} &
\tabparams{690M}{5.9} & 1150.1 &
40.2 & 91.4 & 52.8 & 85.0 & \second{16.4} & 80.3 & 38.7 & 85.0 & 9.1 & 70.9 &
\best{.009} & 96.9 & .095 & 54.1 & .051 & 64.7 & .042 & 69.7 & .311 & 18.4 \\

\addlinespace[2pt]

VGGT~\cite{vggt}\tabvenue{CVPR'25} &
\tabparams{1.26B}{10.8} & 190.0 &
\best{96.5} & \best{99.8} & 35.4 & 82.8 & 11.1 & 79.1 & 15.1 & 74.7 & 39.2 & 83.8 &
.010 & 95.8 & .053 & 52.6 & .042 & 70.8 & .034 & 68.4 & .081 & 42.3 \\

VGGT-$\Omega$~\cite{vggtomega}\tabvenue{CVPR'26} &
\tabparams{1.14B}{9.7} & 99.8 &
77.4 & 98.1 & 64.1 & \second{95.4} & \best{21.3} & \best{87.0} & 29.9 & 87.3 & \second{42.4} & \best{85.5} &
\best{.009} & \second{97.2} & \best{.024} & \best{78.6} & .039 & 65.3 & .032 & 70.6 & \best{.055} & \best{62.3} \\

$\pi^3$~\cite{pi3}\tabvenue{ICLR'26} &
\tabparams{959M}{8.2} & 153.8 &
70.2 & 97.3 & 43.2 & 84.9 & 11.3 & 81.3 & 76.5 & 97.3 & 24.8 & 82.4 &
\best{.009} & \best{97.3} & .034 & 66.8 & \best{.032} & \best{77.8} & \best{.014} & \best{94.3} & .078 & 51.0 \\

MapAnything~\cite{mapanything}\tabvenue{3DV'26} &
\tabparams{1.23B}{10.5} & 148.4 &
18.1 & 88.8 & 37.1 & 73.8 & 8.2 & 74.6 & 70.6 & 96.7 & 37.8 & 82.8 &
.014 & 95.2 & .227 & 40.6 & .044 & 67.9 & .019 & 89.2 & .089 & 51.9 \\

DA3-L~\cite{depthanything3}\tabvenue{ICLR'26} &
\tabparams{411M}{3.5} & 71.4 &
79.4 & 98.4 & 42.0 & 75.6 & 11.4 & 79.7 & 7.0 & 71.8 & 10.9 & 74.6 &
.010 & \third{97.1} & .210 & 49.9 & .039 & 69.7 & .051 & 47.8 & .141 & 27.0 \\

DA3-G~\cite{depthanything3}\tabvenue{ICLR'26} &
\tabparams{1.32B}{11.3} & 178.7 &
\second{86.4} & \second{99.1} & 63.2 & 83.2 & \third{15.4} & 81.3 & 24.3 & 80.1 & 39.8 & 82.3 &
.010 & \third{97.1} & .137 & 64.7 & .037 & 71.0 & .041 & 57.7 & .080 & 42.0 \\

\addlinespace[2pt]

{\dejaview}~\cite{burzio2026dejaview}\tabvenue{NeurIPS'26} &
\tabparams{117M}{} & 75.9 &
83.3 & 98.8 & \second{66.0} & \second{95.4} & 13.9 & 81.7 & \third{79.5} & \third{98.0} & \best{43.4} & \second{85.3} &
\best{.009} & \third{97.1} & \second{.026} & \second{78.3} & \third{.035} & 74.2 & \second{.015} & \second{93.3} & .067 & 58.5 \\

\midrule

\textbf{Solv3R (ours)}, $K{=}16$ &
\tabparams{117M}{} & 76.5 &
84.5 & 99.0 & \third{65.4} & 94.8 & 13.8 & \third{81.8} & \second{80.7} & \best{98.2} & \third{41.5} & \second{85.3} &
.010 & 96.9 & .029 & 74.3 & \third{.035} & \third{74.4} & \third{.016} & 92.5 & \second{.059} & \third{59.9} \\

\textbf{Solv3R (ours)} &
\tabparams{117M}{} & 91.3 &
\third{85.5} & \second{99.1} & \best{70.4} & \best{96.3} & 14.2 & \second{81.9} & \best{81.2} & \best{98.2} & \third{41.5} & \second{85.3} &
.010 & 97.0 & \third{.027} & \third{77.9} & \second{.034} & \second{75.8} & \third{.016} & \third{92.6} & \second{.059} & \second{60.4} \\

\bottomrule
\end{tabular}%
}
\end{center}
\caption{
\textbf{Standard (offline) multi-view reconstruction benchmarks.}
Quantitative comparison with state-of-the-art methods~\cite{mast3r,mast3rsfm,vggt,vggtomega,pi3,mapanything,depthanything3,burzio2026dejaview} across five benchmark datasets~\cite{dtu,eth3d,7scenes,scannetpp,nuscenes}.
We report camera pose AUC@3/@30 ($\uparrow$), geometry relative L2 ($\downarrow$), and inlier ratio (IR, $\uparrow$).
TFLOPs are measured for 24 views.
For our method, $K{=}16$ denotes evaluation after 16 solver calls, while the default uses adaptive stopping based on \cref{eq:relative_update} with $\gamma=0.02$.
}
\label{tab:\currfilebase}
\end{table*}

%% file: fig/incremental_join.tex
\begin{figure*}[tb]
\centering
\includegraphics[width=\linewidth]{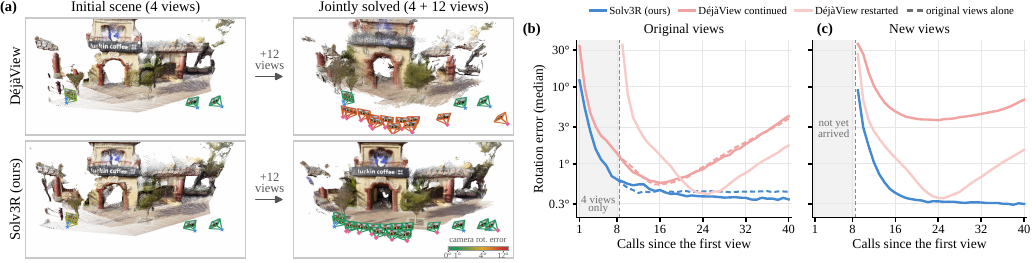}
\caption{\textbf{Revising an existing reconstruction with new views.}
(a) Four views are solved for eight calls before twelve new views are introduced. Camera colors indicate rotation error.
(b,c) Median rotation error of the original and new views during continued refinement over 96 windows.
Dashed curves in (b) indicate the no-join controls, where the original four views continue refinement for the same number of solver calls without introducing new views.
Our method incorporates new views while improving existing camera estimates, whereas \dejaview fails to maintain a consistent reconstruction.
}
\label{fig:\currfilebase}
\end{figure*}

%% file: fig/incremental_efficiency.tex
\begin{figure}[t]
\centering
\includegraphics[width=\linewidth]{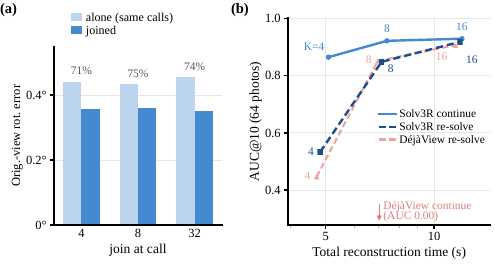}
\caption{\textbf{Revisability and reuse.}
(a) Original-view rotation error when new views join at different refinement stages, compared with the no-join control. Percentages indicate the fraction of improved windows.
(b) Final pose accuracy versus total reconstruction time as the view set expands from 8 to 64.
Solv3R improves existing estimates across different join times and achieves comparable accuracy to re-solving with less computation.}
\label{fig:\currfilebase}
\end{figure}

%% file: sec/5_results.tex
\section{Experiments}
\label{sec:experiments}
We initialize our model from the pretrained \dejaview~\cite{burzio2026dejaview}, with the architectural changes in \cref{sec:geometric_solver}, and train it as in \cref{sec:solver_training} using AdamW~\cite{adamw} with a learning rate of $2{\times}10^{-5}$ and an exponential moving average (EMA) of the weights.
Each training iteration samples 4--64 views and performs 16--40 solver calls without gradients, followed by 8 differentiable calls supervised at the final step.
Before the differentiable calls, we apply feature noise or reset selected views with probability 0.15, and set $\lambda_{\mathrm{dist}}{=}0.5$.
We train for approximately one week on 8$\times$H100 GPUs.
Further implementation details, dataset splits, training configurations, and evaluation protocols are provided in the \SupplementaryMaterial{}.

\subsection{Offline reconstruction}
\label{sec:offline_reconstruction}

We first compare our method with state-of-the-art multi-view reconstruction methods on standard benchmarks.
As shown in \Cref{tab:offline_reconstruction}, Solv3R achieves competitive camera pose and geometry accuracy with a \ul{compact} model.
Importantly, our method already achieves accuracy comparable to \dejaview~\cite{burzio2026dejaview} at the end of its trained 16-call refinement schedule, while exhibiting the stable and corrective solver dynamics demonstrated in \Cref{sec:solver_properties}.
These properties enable continued refinement and reuse of previously computed states beyond a predefined iteration schedule.
The stable solver dynamics also enable adaptive inference using \cref{eq:relative_update}, with $\gamma{=}0.02$ and a maximum of 64 solver calls.
With adaptive stopping, we can further improve most reconstruction metrics over our 16-call inference with only 3--4 \textit{additional} solver calls on average.
Further analyses of stopping behavior and threshold sensitivity are provided in the \SupplementaryMaterial{}.

\subsection{Incremental reconstruction}
\label{sec:incremental_reconstruction}

We next evaluate whether our method can extend and revise an existing reconstruction as new observations arrive.
Using evaluation windows from DL3DV~\cite{dl3dv} and ScanNet++~\cite{scannetpp}, we first solve four views for eight calls, introduce twelve additional views, and continue joint refinement from the existing states.
As shown in \Cref{fig:incremental_join}, our method reconstructs the new views while improving the camera estimates of the original views, whereas \dejaview fails to maintain a consistent reconstruction.
Compared with the no-join control using the same number of solver calls, this improvement shows that new observations contribute to correcting the existing solution rather than merely benefiting from further computation.
In \Cref{fig:incremental_efficiency}, we examine revisability across different refinement stages and the efficiency of \textit{reusing} previous computation.
Adding new views at different refinement stages, even after the initial reconstruction meets the stopping criterion, improves original-view accuracy in most evaluation windows.
When incrementally expanding the reconstruction from 8 to 64 views, our method achieves accuracy comparable to re-solving from scratch with fewer solver calls and lower total runtime.
These results show that existing reconstructions can be revised as new observations arrive while reusing previous computation, rather than repeatedly solving from scratch.

\input{fig/reloc}
\input{tab/relocalization}

\subsection{Map-based relocalization}
\label{sec:relocalization}

We evaluate map-based relocalization on 7-Scenes~\cite{7scenes} and Cambridge~\cite{cambridge}, using maps reconstructed from the training frames of each scene.
For each query, we freeze the solved map representations and refine only the query.
We select the 10 nearest map views of each query with an off-the-shelf image retrieval model~\cite{netvlad}.
The query attends to the cached states of these views and of map views that cover the scene.
All methods attend to the same map views for each query.
This setting tests whether previously solved geometry can serve as reusable context for subsequent inference.
Conventional GFMs require layer-specific KV caches to preserve geometric context.
Although \dejaview shares a single transformer block across iterations, reusing only the KV cache of its solved map fails to relocalize queries, requiring separate caches for each refinement stage.
In contrast, Solv3R successfully reuses a single solved-map KV cache across all solver calls.
As shown in \Cref{tab:relocalization}, our method achieves the highest localization recall on both benchmarks with lower cache storage than layer-wise or stage-specific caching alternatives.
\Cref{fig:reloc_map} shows the queries of one scene, where \dejaview with a single cache scatters them while Solv3R places them where they belong.
These results demonstrate that our method produces geometric representations that remain reusable across solver calls, enabling subsequent inference without stage-specific map computation.

\input{fig/streaming}
\input{tab/streaming}
\subsection{Streaming reconstruction}
\label{sec:streaming_reconstruction}

We evaluate Solv3R against state-of-the-art streaming reconstruction methods across benchmarks covering camera pose, video depth, and 3D reconstruction.
We do not introduce streaming-specific training or a separate optimization backend.
Instead, we leverage the solver properties established in \Cref{sec:solver_properties} to maintain and update an ongoing reconstruction as new frames arrive.
An initial set of anchor views establishes a shared reference frame, while incoming views are warm-started and jointly refined with recent observations (\cref{prop:continuation}).
Previously solved views provide frozen, KV-cached geometric context (\cref{prop:stability}), but can be reactivated for joint refinement when new observations reveal inconsistencies (\cref{prop:correction}).
This allows the reconstruction to incorporate new information and correct earlier estimates without restarting or chaining separate reconstructions.
Details of window management, memory selection, and cache refresh are provided in the \SupplementaryMaterial{}.
\input{fig/stream_bubble}
Although incoming frames receive further refinement while remaining in the active window, we report predictions at frame arrival for fair comparison with existing streaming methods.

As shown in \Cref{tab:streaming}, Solv3R achieves competitive accuracy and efficiency across these tasks using only its solver properties.
\Cref{fig:streaming} further illustrates accurate camera trajectories over long sequences, while \Cref{fig:streaming_bubble} shows the accuracy--efficiency trade-off.
Further ablations in the \SupplementaryMaterial{} show that retaining previously solved states and periodically reactivating frozen views are important for maintaining reconstruction accuracy over long sequences.
These results demonstrate that the same learned solver can achieve competitive streaming reconstruction by continuing and revising existing solutions, without streaming-specific training or a separate optimization backend.

\subsection{Ablation studies}
\label{sec:ablation}
We examine the contributions of individual components to the solver dynamics in \Cref{tab:ablation}.
For controlled ablations, we fine-tune each configuration for 3k steps from the same checkpoint~\cite{burzio2026dejaview}.
While reconstruction accuracy remains comparable at 16 calls, removing individual components reveals distinct failure modes under continued computation and state reuse.
Step-conditioned gates also benefit from our training, but still depend on the planned number of calls $P$, which is not suitable for continued solving.
These variants further require a separate KV cache for every step, yet remain less accurate than our single-cache model.
Removing input injection, state normalization, or random-depth pre-calls causes accuracy to collapse at 1024 calls.
In contrast, removing perturbation training largely preserves long-horizon accuracy but degrades joining and cached-context inference, showing that stability alone does not guarantee corrective refinement.
With sufficient training, Solv3R also recovers the initial drop in fixed-budget accuracy while retaining stable and corrective solver dynamics.
Together, these results demonstrate that accurate fixed-budget predictions are insufficient for reusable geometric solving, which requires both stable continuation and corrective updates.

\input{tab/ablation}

%% file: fig/reloc.tex
\begin{figure*}[t]
\centering
\includegraphics[width=\linewidth]{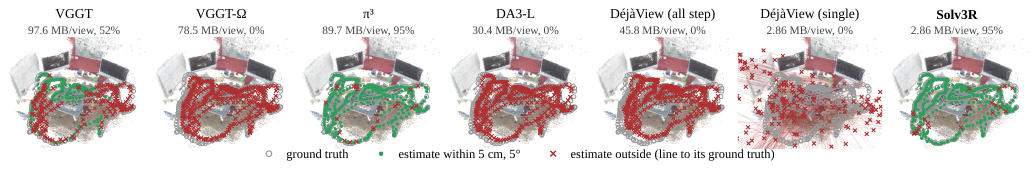}
\caption{\textbf{Relocalization against a frozen map.}
Green dots are query estimates within 5\,cm and 5$^\circ$ of the ground truth (gray rings), and red crosses are the others, linked to their ground truth.
Above each panel are the cache size per map view and the fraction of queries within 5\,cm and 5$^\circ$.
With the smallest cache, our model localizes as many queries as the most accurate baseline.}
\label{fig:reloc_map}
\end{figure*}

%% file: tab/relocalization.tex
\begin{table}[t]
\begin{center}
\resizebox{\linewidth}{!}{%
\setlength{\tabcolsep}{2pt}%
\setlength{\fboxsep}{1pt}%
\renewcommand{\arraystretch}{1.08}%
\fontsize{6.5}{8}\selectfont
\begin{tabular}{@{}lcccccc}
\toprule
& \multirow{2}{*}{\shortstack{Cache\\(MB/view)$\downarrow$}}
& \multirow{2}{*}{\shortstack{Time\\(ms)$\downarrow$}}
& \multicolumn{2}{c}{\textbf{7-Scenes}}
& \multicolumn{2}{c}{\textbf{Cambridge}} \\
\cmidrule(lr){4-5}
\cmidrule(lr){6-7}
\textbf{Method}
& &
& Med.$\downarrow$ & 5cm/5$^\circ\uparrow$
& Med.$\downarrow$ & 25cm/2$^\circ\uparrow$ \\
\midrule
VGGT & 97.6\phantom{0} & 186 & \third{6.35 / 1.35} & \third{45.5\nocorr} & 183.2 / 1.77 & \second{16.4\nocorr} \\
VGGT-$\Omega$ & 78.5\phantom{0} & 158 & 8.87 / 1.68 & 32.8 & 187.7 / 1.18 & \third{14.9\nocorr} \\
$\pi^3$ & 89.7\phantom{0} & 161 & \second{4.11 / 0.99} & \second{66.9\nocorr} & \second{85.9 / 0.95} & 14.7 \\
DA3-L & \third{30.4\nocorr}\phantom{0} & \phantom{0}\third{62\nocorr} & 6.90 / 1.16 & 36.7 & 147.9 / 1.86 & 12.7 \\
\addlinespace[2pt]
\dejaview (all step) & 45.8\phantom{0} & \phantom{0}88 & 11.30 / 1.31 & 28.3 & \third{131.0 / 1.07} & \phantom{0}0.7 \\
\dejaview (single) & \phantom{0}\best{2.86} & \phantom{0}\best{50} & 123.9 / 29.9 & \phantom{0}0.0 & 3615 / 51.5 & \phantom{0}0.0 \\
\midrule
\textbf{Solv3R (ours)} & \phantom{0}\best{2.86} & \phantom{0}\second{51\nocorr} & \best{3.58 / 0.95} & \best{72.8} & \best{63.5 / 0.74} & \best{24.9} \\
\bottomrule
\end{tabular}%
}
\end{center}
\caption{\textbf{Map-based relocalization.}
Evaluation on indoor 7-Scenes~\cite{7scenes} and outdoor Cambridge Landmarks~\cite{cambridge}.
Median translation/rotation errors (cm/$^\circ$), localization recall (\%), cache size per view (MB), and query latency (ms).}
\label{tab:\currfilebase}
\end{table}

%% file: fig/streaming.tex
\begin{figure*}[t]
\centering
\includegraphics[width=0.95\linewidth]{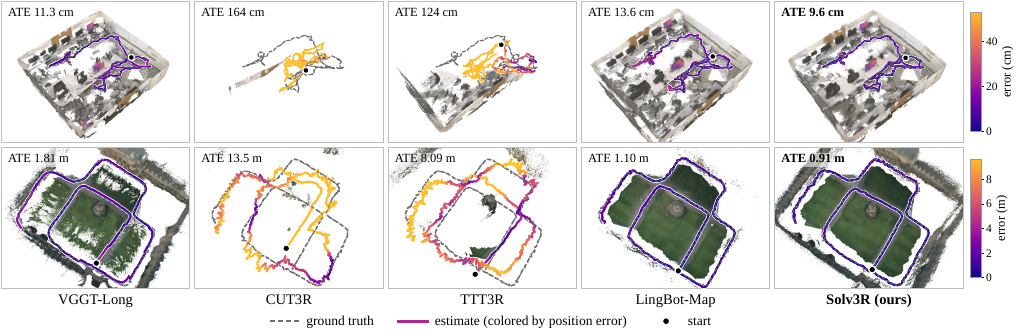}
\caption{\textbf{Streaming reconstruction.}
Qualitative comparison with window-based reconstruction~\cite{vggtlong}, recurrent-state methods~\cite{cut3r,ttt3r}, and geometric-context approaches~\cite{lingbotmap} on long sequences, with camera trajectories colored by position error.}
\label{fig:\currfilebase}
\end{figure*}

%% file: tab/streaming.tex
\providecommand{\tabvenue}[1]{%
  \hspace{2pt}{\color{gray}\fontsize{5.7}{6.8}\selectfont (#1)}%
}

\begin{table*}[tb]
\begin{center}
\resizebox{\textwidth}{!}{%
\setlength{\tabcolsep}{4pt}%
\setlength{\fboxsep}{1pt}%
\renewcommand{\arraystretch}{1.13}%
\fontsize{7}{8.5}\selectfont
\begin{tabular}{@{}l*{15}{c}}
\toprule
\multirow{3}{*}{\textbf{Method}}
& \multicolumn{2}{c}{\textbf{Efficiency}}
& \multicolumn{3}{c}{\textbf{Camera Pose}}
& \multicolumn{6}{c}{\textbf{Video Depth}}
& \multicolumn{4}{c}{\textbf{3D Reconstruction}} \\
\cmidrule(lr){2-3} \cmidrule(lr){4-6}
\cmidrule(lr){7-12} \cmidrule(l){13-16}
& Time & Mem.
& Sintel & TUM-dyn. & ScanNet
& \multicolumn{2}{c}{Sintel}
& \multicolumn{2}{c}{Bonn}
& \multicolumn{2}{c}{KITTI}
& \multicolumn{2}{c}{7-Scenes}
& \multicolumn{2}{c}{NRGBD} \\
\cmidrule(lr){7-8} \cmidrule(lr){9-10}
\cmidrule(lr){11-12} \cmidrule(lr){13-14}
\cmidrule(l){15-16}
& (ms)$\downarrow$ & (GB)$\downarrow$
& ATE$\downarrow$ & ATE$\downarrow$ & ATE$\downarrow$
& AbsRel$\downarrow$ & $\delta\uparrow$
& AbsRel$\downarrow$ & $\delta\uparrow$
& AbsRel$\downarrow$ & $\delta\uparrow$
& Cham.$\downarrow$ & NC$\uparrow$
& Cham.$\downarrow$ & NC$\uparrow$ \\
\midrule

\addlinespace[2pt]
\multicolumn{16}{@{}l}{%
    \scriptsize\color{black!55}\itshape w/ Separate back-end
} \\

DROID-SLAM$^\dagger$~\cite{droidslam}\tabvenue{NeurIPS'21}
& \phantom{0}\best{36} & \phantom{0}5.7
& .164 & .065 & .137
& -- & -- & -- & -- & -- & --
& -- & -- & -- & -- \\

MASt3R-SLAM~\cite{mast3rslam}\tabvenue{CVPR'25}
& \phantom{0}73 & \phantom{0}7.2
& .273 & .102 & .139
& -- & -- & -- & -- & -- & --
& -- & -- & -- & -- \\

SLAM-Former~\cite{slamformer}\tabvenue{ECCV'26}
& \phantom{0}72 & 10.4
& \best{.147} & \third{.040} & .096
& -- & -- & -- & -- & -- & --
& -- & -- & -- & -- \\

VGGT-Long$^\ddagger$~\cite{vggtlong}\tabvenue{ICRA'26}
& 584 & 19.9
& .172 & .187 & \third{.090}
& \best{.302} & \best{.681}
& \third{.068} & \second{.957}
& .120 & .867
& \third{.022} & \best{.592}
& \best{.009} & \best{.765} \\

\addlinespace[3pt]
\multicolumn{16}{@{}l}{%
    \scriptsize\color{black!55}\itshape Online reconstruction
} \\

CUT3R~\cite{cut3r}\tabvenue{CVPR'25}
& \phantom{0}\second{52\nocorr} & \phantom{0}6.1
& .204 & .170 & .843
& .431 & .471
& .084 & .939
& .130 & .866
& .069 & .566
& .085 & .616 \\

TTT3R~\cite{ttt3r}\tabvenue{ICLR'26}
& \phantom{0}\second{52\nocorr} & \phantom{0}6.1
& .200 & .112 & .416
& .406 & .487
& .076 & .953
& .118 & .893
& .026 & .582
& .038 & \third{.634} \\

Spann3R~\cite{spann3r}\tabvenue{3DV'25}
& \phantom{0}78 & \phantom{0}\best{2.6}
& .311 & .147 & .767
& .649 & .325
& .091 & .930
& .343 & .442
& .038 & .566
& .046 & .615 \\

Point3R~\cite{point3r}\tabvenue{NeurIPS'25}
& 192 & \phantom{0}\second{3.5\nocorr}
& .388 & .171 & .694
& .478 & .452
& .082 & .948
& \third{.104} & \third{.915}
& .034 & .578
& .054 & .621 \\

LingBot-Map~\cite{lingbotmap}\tabvenue{ECCV'26}
& 119 & 14.3
& \second{.149} & \second{.032} & \second{.089}
& \second{.322} & \third{.648}
& \second{.067} & \best{.959}
& \best{.070} & \best{.955}
& \second{.021} & \third{.586}
& \second{.016} & .633 \\

\midrule

\textbf{Solv3R (ours)}
& \phantom{0}95 & \phantom{0}\third{5.0\nocorr}
& \third{.157} & \best{.015} & \best{.077}
& \second{.322} & \second{.668}
& \best{.064} & \third{.956}
& \second{.078} & \second{.947}
& \best{.018} & \second{.591}
& \third{.018} & \second{.640} \\

\bottomrule
\end{tabular}%
}

{\raggedright\fontsize{7}{8}\selectfont
$^\dagger$ GT camera intrinsics.
$^\ddagger$ Offline loop closure.
``--'' indicates dense metrics not evaluated for keyframe-based methods.\par}
\end{center}
\caption{
\textbf{Streaming reconstruction benchmarks.}
Comparison with state-of-the-art streaming reconstruction methods across camera pose, video depth, and 3D reconstruction.
We report per-frame runtime (ms) and peak GPU memory (GB), together with camera trajectory ATE (m), depth AbsRel and $\delta_{1.25}$, and reconstruction Chamfer distance (m) and normal consistency (NC).
}
\label{tab:\currfilebase}
\end{table*}

%% file: fig/stream_bubble.tex
\begin{figure}[t]
\centering
\includegraphics[width=0.95\linewidth]{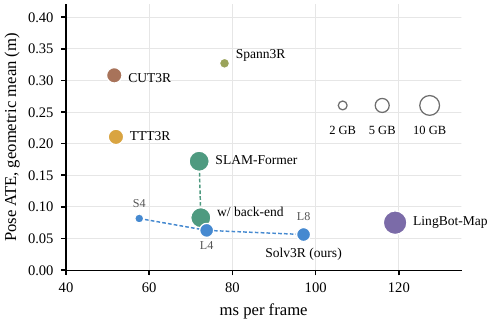}
\caption{\textbf{Streaming accuracy vs. efficiency.}
Comparison of online reconstruction methods in terms of geometric-mean pose ATE and per-frame runtime. Bubble size indicates peak GPU memory.
For Solv3R, S/L denotes small/large memory configurations, and numbers indicate solver calls per frame (L8 is the default).}
\label{fig:streaming_bubble}
\end{figure}

%% file: tab/ablation.tex
\begin{table}[t]
\begin{center}
\resizebox{\linewidth}{!}{%
\setlength{\tabcolsep}{3pt}%
\renewcommand{\arraystretch}{1.1}%
\fontsize{7.5}{9}\selectfont
\begin{tabular}{@{}lcccccc@{}}
\toprule
& \multicolumn{3}{c}{AUC@5$\uparrow$} & Recovery$\downarrow$ & Join$\downarrow$ & Cache$\downarrow$ \\
\cmidrule(lr){2-4}
\textbf{Method} & 16 & 128 & 1024 & ($^\circ$) & ($^\circ$) & ($^\circ$) \\
\midrule
\dejaview{} (released) & 76.5 & \phantom{0}0.0 & \phantom{0}0.0 & 8.37 & 1.80 & 55.5$^\dagger$ \\
\midrule
\multicolumn{7}{@{}l}{\textit{Fine-tuned for 3k steps}} \\
\dejaview{} w/ our training & 73.5 & 19.3 & \phantom{0}0.0 & 3.94 & 1.31 & \phantom{0}1.05$^\dagger$ \\
Solv3R & 72.5 & 70.8 & 65.4 & 1.26 & 0.62 & \phantom{0}0.76 \\
\quad w/ step-conditioned ($P{=}16$) & 71.5 & 52.1 & 23.5 & 1.96 & 0.81 & \phantom{0}0.95$^\dagger$ \\
\quad w/ step-conditioned ($P{=}1024$) & 61.9 & 43.3 & 52.2 & 4.33 & 2.53 & \phantom{0}2.58$^\dagger$ \\
\quad w/o input injection & 71.6 & 33.1 & \phantom{0}0.0 & 5.98 & 0.64 & \phantom{0}0.74 \\
\quad w/o state normalization & 74.3 & 70.5 & \phantom{0}0.0 & 0.94 & 0.59 & \phantom{0}0.72 \\
\quad w/o random-depth pre-calls & 73.1 & \phantom{0}3.4 & \phantom{0}0.0 & 7.30 & 4.37 & \phantom{0}4.15 \\
\quad w/o perturbation & 71.0 & 69.1 & 61.7 & 3.32 & 4.24 & \phantom{0}3.51 \\
\midrule
\textbf{Solv3R (full method)} & 77.2 & 79.1 & 78.8 & 0.22 & 0.54 & \phantom{0}0.72 \\
\bottomrule
\end{tabular}%
}

{\raggedright\fontsize{8}{9.5}\selectfont
$^\dagger$ All-step KV caches (one per step). Other rows use a single cache.\par}
\end{center}
\caption{\textbf{Ablation study.}
Solv3R and its variants are fine-tuned for 3k steps from the same initialization, with \dejaview and our full method as references.
We report pose accuracy at different solver calls, along with rotation errors after perturbation (Recovery), new-view insertion (Join), and inference with frozen KV context (Cache).
For step-conditioned gates, $P$ is the number of calls their schedule is planned for.}
\label{tab:\currfilebase}
\end{table}

%% file: sec/6_conclusion.tex
\section{Conclusion}
\label{conclusion}
In this paper, we presented Solv3R, which reformulates multi-view 3D reconstruction as an ongoing geometric solving process.
We identified the properties required for continued solving and proposed a stationary update operator and a training scheme to realize them in a looped transformer.
Our learned solver can \textit{continue} refining existing estimates, \textit{integrate} new observations, and \textit{reuse} previously computed geometric representations, allowing a single model to address diverse reconstruction settings that are typically handled by separate architectures.
This shifts the role of a geometric foundation model from fixed-set prediction to reusable geometric computation.
We believe this opens a path toward a more open-ended paradigm of 3D reconstruction, where geometric estimates can be refined and extended from their current states as observations change.
More broadly, this perspective could extend to vision-language models, robotics, and world modeling, where maintaining and refining evolving states is a fundamental challenge.

%% file: sec/X_suppl.tex
\clearpage
\appendix

\section{Implementation details}
\label{sec:supp_implementation}

We provide details of the architecture, training, and streaming
configuration of Solv3R.

\paragraph{Architecture.}
We build on the pretrained \dejaview~\cite{burzio2026dejaview}, retaining its shared transformer block with frame-wise and global attention.
We replace the step-conditioned gates with learnable constant gates, which provide the same channel-wise scaling in each sub-block across all solver calls.
We initialize these gates by averaging the corresponding gate values over the 16 refinement steps of \dejaview.

To keep the state grounded in the original observations and stable under continued computation, we inject the initial state at every solver call and normalize the updated state:
\begin{equation}
    \mathbf{u} = \mathcal{U}_\theta(\mathbf{z}^k + W\bar{\mathbf{z}}^0),
    \qquad
    \mathbf{z}^{k+1} = \frac{\rho\,\mathbf{u}}{\|\mathbf{u}\|_{\mathrm{RMS}}}.
\end{equation}
Here, $\mathbf{z}^k$ is the state after $k$ solver calls, $\mathcal{U}_\theta$ denotes the shared block, $\bar{\mathbf{z}}^0$ is the layer-normalized~\cite{layernorm} initial state without affine parameters, and $W$ is a zero-initialized linear projection.
The target RMS $\rho$ is the RMS of the final \dejaview state, measured once before training ($\rho \approx 0.85$), and stays fixed throughout training and inference.
We scale the entire state by a single factor, which preserves relative feature magnitudes across views and tokens.
The resulting model has 117M parameters.
It takes images whose longer side is 504 pixels.

\paragraph{Training data.}
To cover diverse scenes, we train on a broad mixture of indoor, object-centric, outdoor, driving, and synthetic datasets, summarized in \Cref{tab:supp_training_data}.

\input{tab/supp_training_data}

\paragraph{Training objective.}
The geometric loss $\mathcal{L}_{\mathrm{geo}}$ supervises the ray maps, cameras, and depths decoded at the final call.
For ray maps, we apply an L1 loss to ray directions and scale-aligned ray origins.
For cameras, we use geodesic losses on the relative rotations and translation directions.
For depth, we use a scale-aligned log-L1 loss with weight 0.5.
On datasets with dense and accurate depth, we add a multi-scale depth-gradient loss~\cite{megadepth} with weight 0.5.
These terms align each prediction to the ground truth with its own scale.
They do not enforce a common scale across views or between depths and camera translations, so we add two consistency terms with weight 1.5 each.
The first penalizes the deviation of each view's depth scale from the window mean.
The second supervises the camera centers and ray origins after rescaling them by the window's depth scale.
The distillation loss $\mathcal{L}_{\mathrm{dist}}$ applies the ray, log-depth, and camera losses against the 16-step predictions of \dejaview, with $\lambda_{\mathrm{dist}} = 0.5$.
For datasets whose ground truth is sparse or noisy, marked with $^\ast$ in \Cref{tab:supp_training_data}, we use $\mathcal{L}_{\mathrm{dist}}$ only.

\paragraph{Training procedure.}
To cover different window sizes, we sample each window from one of three size classes: 4--16, 24--32, or 48--64 views, with probabilities 0.4, 0.3, and 0.3.
Starting from $\mathbf{z}^0$, we apply a number of solver calls without gradients, drawn uniformly from 16 to 40.
We then apply eight differentiable calls and supervise the final call.
With probability 0.15, we perturb the state before the differentiable calls.
We either add Gaussian noise with a standard deviation of $\sigma\,\|\mathbf{z}\|_{\mathrm{RMS}}$, where $\sigma$ is drawn uniformly from $[0, 0.5]$, or reset a random subset of up to half of the views to $\mathbf{z}^0$.
The noise exposes the solver to globally inaccurate states.
The reset mixes refined and unrefined views within one window, as occurs when views are inserted into an existing reconstruction.

\paragraph{Optimization.}
We train in two stages.
The first stage gradually increases the window size and the number of solver calls.
It runs for about 70k iterations with a batch of 8 windows.
The second stage trains with the full procedure above for about 20k iterations with a batch of 32 windows.
To cover rolled cameras, we rotate each view by $90^\circ$ in either direction with probability 0.15 in its last 2k iterations.
We fine-tune only the shared block, the gates, and the projection $W$, and freeze the image encoder and the prediction heads of \dejaview.
We use AdamW~\cite{adamw} with a weight decay of 0.01 and learning rates of $2\times10^{-5}$ for the shared block and the projection, and $2\times10^{-4}$ for the gates.
We evaluate an exponential moving average of the weights with a decay of 0.999.

\paragraph{Streaming reconstruction.}
Streaming uses the final model without any streaming-specific training.
For each incoming frame, the solver operates on a window of anchors, memory frames, recent frames, and the new frame.
The first $a$ frames serve as anchors that fix the coordinate frame and are initially solved jointly for 16 calls.
The $r$ most recent frames form the active window, where each new frame is initialized from its image encoding plus the solved offset of the previous frame.
The solver then jointly updates the new and recent frames for $m$ calls, retaining their previously computed states (\textbf{continuation}).
Anchors and memory frames remain frozen during normal solver calls, providing fixed attention context.
Since their representations remain unchanged, their key--value (KV) pairs are computed once and reused across solver calls (\textbf{stability}).
When a frame leaves the recent window, it is retained in memory if its index is a multiple of $s$, and discarded otherwise.
The memory holds the latest $M$ such frames.
To detect inconsistencies with newly observed frames, we periodically check whether the frozen states would change under further refinement.
Every four incoming frames, one additional solver call evaluates the relative state update $\|\mathbf{z}^{k+1}-\mathbf{z}^{k}\|_2/\|\mathbf{z}^{k}\|_2$ on the frozen views.
If this exceeds $\gamma=0.02$, the frozen views rejoin joint refinement with the incoming frame, continuing from their existing states rather than restarting from scratch (\textbf{correction}).
Their KV cache is then rebuilt from the updated states; otherwise, the cached representations are retained.
The default configuration (L8) uses $a=8$, $r=6$, $s=16$, $M=64$, and $m=8$.

\section{Evaluation protocols}
\label{sec:supp_protocols}

We detail the evaluation protocols for the solver-property analyses and the offline, incremental, map-based relocalization, and streaming experiments.

\paragraph{Solver properties.}
The stability analysis of the main paper uses 148 eight-view windows, 66 from DTU~\cite{dtu}, 37 from ETH3D~\cite{eth3d}, and 45 from ScanNet++~\cite{scannetpp} scenes held out from our fine-tuning.
We measure rotation errors relative to the first view of each window.
To test recovery, we perturb the state of 36 of these windows, 12 from each dataset, at call 8 and report the rotation error at call 16.
Relative noise (rel.) is Gaussian noise with a standard deviation of one or two times the RMS of the state of each model.
Fixed-scale noise (fixed) uses the mean state RMS of our model and \dejaview instead, so that both models receive the same absolute noise.
A reset returns every view except the first to its initial state.
Apart from the KITTI sequence in \cref{fig:warm_start} (left), the other solver-property experiments use 64-frame sequences from held-out DL3DV~\cite{dl3dv} scenes and the held-out ScanNet++ scenes.

Warm start simulates a video that arrives one frame at a time.
The window holds the first frame, which fixes the coordinate frame, and the 23 most recent frames.
When a new frame arrives, the frames that stay in the window keep their solved states.
The new frame starts from its initial state plus the solved offset of the previous frame, as in streaming (\cref{sec:supp_implementation}).
Over 16 sequences from each dataset, we count the solver calls until the rotation error of the new frame falls below 2$^\circ$, and compare them with solving the window from scratch.

Pipelined reconstruction slides a window of the first frame and the 16 most recent frames over a sequence.
It advances the window after every solver call rather than after convergence.
After each call, the oldest frame other than the first leaves the window, and the next frame enters from its initial state.
Apart from the first frame, each call thus updates frames that have received between 0 and 15 previous calls.
We evaluate the pose of each frame when it leaves the window, after its 16th call.

Random computational histories test whether the joint update depends on how each view reached its current state.
We record the states of a window over its first 16 calls from scratch, and build a new state by taking each view from a randomly chosen call of this trajectory.
We then run 16 joint calls from this state, and report the mean difference of the camera rotations across 20 such draws for each of 96 windows of 16 views.

We evaluate whether additional geometric constraints can correct an inaccurate reconstruction without restarting the solve.
We construct 39 weak-overlap windows, 9 from ETH3D~\cite{eth3d} and 30 from Mill19~\cite{meganerf}, each containing two groups of four views with at most 5\% co-visibility.
We select the 17 windows with inaccurate initial reconstructions, whose median rotation errors are $7.3^\circ$ for our model and $8.4^\circ$ for \dejaview.
We then add 8 or 16 bridging views that observe both groups.
We compare continuing from the stopped states for 64 additional calls with re-solving all views from their initial states under the same total number of solver calls.

We evaluate whether partial updates can approach the accuracy of joint reconstruction.
We use 20 scenes of 128 views from 7-Scenes~\cite{7scenes}, DL3DV~\cite{dl3dv}, ScanNet++~\cite{scannetpp}, and Tanks and Temples~\cite{tanksandtemples}.
At each solver call, eight randomly selected views are updated, while the remaining views provide fixed KV-cached context.
We compare against joint solving at the same average number of updates per view, over three random block orders.

\paragraph{Offline reconstruction.}
We follow the evaluation protocol and splits of \dejaview~\cite{burzio2026dejaview} on DTU~\cite{dtu}, ETH3D~\cite{eth3d}, 7-Scenes~\cite{7scenes}, ScanNet++~\cite{scannetpp}, and nuScenes~\cite{nuscenes}.
We report the area under the curve (AUC) of relative pose errors at 3$^\circ$ and 30$^\circ$, and the relative L2 error and inlier ratio of the world-frame point maps.
For adaptive inference, we stop when the relative update falls below $\gamma=0.02$, with at least 4 and at most 64 solver calls.
Without adaptive stopping, our model runs 16 solver calls, the budget of \dejaview.

\paragraph{Incremental reconstruction.}
We use 20 scenes, 10 from the held-out DL3DV scenes and 10 from the held-out ScanNet++ scenes.
Starting from 8 views in capture order, we add 8 views at a time up to 64 views, and report the pose AUC@10 over all view pairs after each insertion.

\paragraph{Map-based relocalization.}
For each scene, we solve the training frames once as a map and cache the key-value (KV) pairs of every map frame.
For 7-Scenes~\cite{7scenes}, we use every fifth training frame to construct the map and every fifth test frame as a query.
For Cambridge~\cite{cambridge}, we use all test frames as queries.
Each query is refined for 16 solver calls against the fixed KV-cached map representations.
The attended frames are its 10 nearest map frames retrieved by NetVLAD~\cite{netvlad}, the reference view of the map, and 64 evenly spaced map frames that cover the scene.
As in map-based relocalization~\cite{reloc3r}, the poses of the map frames are known.
Since each method predicts poses in its own coordinate frame, we recover the absolute query pose by aligning the predicted poses of the retrieved frames to their ground-truth poses with a similarity transform.
We report the median translation and rotation errors and the recall within 5\,cm and 5$^\circ$ for 7-Scenes~\cite{7scenes} and within 25\,cm and 2$^\circ$ for Cambridge~\cite{cambridge}.
All methods attend to the same map frames and use the same alignment.
We time each query on one H100 GPU, from encoding the query to reading out its pose, including copying the cached key-value pairs of its retrieved frames from host memory.

\paragraph{Streaming reconstruction.}
We evaluate on Sintel~\cite{sintel}, TUM dynamics~\cite{tumrgbd}, ScanNet~\cite{scannet}, Bonn~\cite{bonn}, KITTI~\cite{kitti}, 7-Scenes~\cite{7scenes}, and NRGBD~\cite{nrgbd}.
We use 1000 frames with a stride of 3 for ScanNet, 500 frames for Bonn and KITTI, and 200 frames with a stride of 2 for 7-Scenes and NRGBD.
We evaluate each frame with its prediction at arrival, before any later refinement.
The first eight frames are an exception, since we solve them jointly as anchors.
We report the absolute trajectory error (ATE) after Sim(3) alignment~\cite{umeyama}, the depth AbsRel and $\delta_{1.25}$ after per-sequence scale alignment, and the Chamfer distance and normal consistency of the reconstructed point clouds.

\paragraph{Ablation study.}
We evaluate the ablated models in terms of stability under continued computation, recovery from perturbations, new-view insertion, and inference with cached geometric context (\cref{tab:ablation}).
Stability is measured by pose AUC@5 after 16, 128, and 1024 solver calls on the 148 windows used for the stability analysis.
For recovery, we perturb all but the reference view of a 16-call solution with Gaussian noise of standard deviation equal to the state RMS, and measure the median rotation difference from the unperturbed solution after 16 additional calls.
For new-view insertion, we add 12 views to four views solved for 16 calls and report the new-view rotation error after 16 joint calls over 60 windows from DTU and ETH3D.
For cached-context inference, we refine the last view of each of the 148 windows against the cached key--value pairs of the other seven solved views.

\section{Additional analyses}
\label{sec:supp_analyses}

We further examine the properties that distinguish a learned solver from a looped model with a fixed schedule.
In particular, we investigate whether solved states remain accurate under continued computation, can be reused as fixed context, can be revised by new constraints, and can be refined jointly or selectively, regardless of how many calls each view has received.
We also analyze the attention patterns of the solver and its limitations.

\subsection{Stopping and initialization}
\label{sec:supp_convergence}

\paragraph{Stopping behavior.}
As state updates diminish, geometric error decreases and levels off by the stopping point, remaining low under further refinement (\cref{fig:stability}).
Adaptive stopping matches the median accuracy of continuing to 64 calls.
Of the 148 windows of the stability analysis, 144 reach $\gamma=0.02$ within 64 calls, after 17.5 calls on average (median 17, interquartile range 15--19).
Over these windows, the median rotation error is $0.44^\circ$ at the stopping call and $0.43^\circ$ at call 64.
These results show that state-based stopping closely matches the geometric accuracy obtained with further refinement across the evaluated windows.
The other four windows are also accurate at call 64, with rotation errors below $3^\circ$.
\dejaview does not reach the threshold in any window.

\paragraph{Distribution of stopping calls.}
The number of calls adapts to each window (\cref{fig:supp_stopdist}).
All 144 windows stop between 13 and 32 calls.
Our solver runs more calls on harder windows.
The stopping call correlates with the rotation error at that call, with a Spearman correlation of 0.77.
The median stopping call also differs across datasets, from 14 calls on ScanNet++ to 18.5 calls on DTU.

\input{fig/supp_stopdist}

\paragraph{Sensitivity to the threshold.}
The threshold $\gamma=0.02$ balances accuracy and cost (\cref{tab:supp_tau}).
A tighter threshold needs 1.6 to 2.7 times as many calls without consistent gains.
A looser threshold saves about a third of the calls but loses 4 to 8 AUC points on DTU, ETH3D, and ScanNet++.

\input{tab/supp_tau}

\paragraph{Other datasets.}
Stopping remains nearly as accurate as continuing to 128 calls on datasets outside the stability analysis (\cref{tab:supp_hard}).
We solve 107 more eight-view windows from Tanks and Temples~\cite{tanksandtemples}, 7-Scenes~\cite{7scenes}, and TUM dynamics~\cite{tumrgbd} with moving people.
Our model reaches $\gamma=0.02$ in 105 of these windows.
On each dataset, its median rotation error at the stopping call is within $0.04^\circ$ of that at call 128.
No window stops with an error above $5^\circ$.
\dejaview reaches the threshold in none of the windows.

\paragraph{Different initial states.}
The solver reaches nearly the same camera rotations from different initial states.
We solve each window again from two other initial states.
The noisy start adds Gaussian noise to $\mathbf{z}^0$ with a standard deviation of half the RMS of each token.
In the swapped start, each view except the first starts from the initial state of another view, which is a valid state of a different camera.
After 128 calls, the median rotation difference between the start from $\mathbf{z}^0$ and either other start is at most $0.11^\circ$ in every row of \cref{tab:supp_hard}.

\input{tab/supp_hard}

\paragraph{Initialization of new views.}
The solver converges from different initial states of a new view.
We repeat the warm-start experiment of the main paper with three initializations of the new frame (\cref{fig:supp_warminit}).
The new frame starts from its own initial state $\mathbf{z}^0$, from $\mathbf{z}^0$ plus the solved offset of the previous frame, or from the solved state of the previous frame.
The offset gives the most accurate first call, but $\mathbf{z}^0$ reaches the same error within two to four calls.
Even when the new frame starts from the solved state of the previous frame, which describes a different camera, the solver corrects it within about 10 calls.
All three initializations converge to the error of solving the whole window from scratch.

The initialization matters more when each frame receives only a few calls, as in streaming.
We run the default streaming window with 1 to 16 calls per frame (\cref{fig:supp_streamwarm}).
We compare a cold start from $\mathbf{z}^0$ with a warm start from $\mathbf{z}^0$ plus the solved offset of the previous frame.
With one call per frame, the warm start reduces the ATE about sixfold on TUM dynamics and 3.5-fold on 7-Scenes.
From eight calls per frame, both initializations give nearly the same accuracy.
We thus warm-start new frames in all streaming configurations.

\input{fig/supp_warminit}

\input{fig/supp_streamwarm}

\subsection{Streaming memory and long sequences}
\label{sec:supp_streaming}

\paragraph{Configurations.}
The number of calls per frame and the window size set the speed of streaming, and the window size also sets its GPU memory.
We compare the default configuration (L8) with L4, which halves the calls per frame, and with S4, which also uses fewer anchors, recent frames, and memory frames (\cref{tab:supp_stream_cfg}).
L4 shortens the runtime by about a third while keeping its video depth nearly unchanged.
S4 needs 45\% less runtime and less than half the GPU memory of L8.
Compared with CUT3R~\cite{cut3r} and TTT3R~\cite{ttt3r} in the streaming benchmark of the main paper, S4 remains more accurate in camera pose and video depth, at a similar runtime and a third of their GPU memory.

\paragraph{On-arrival and refined predictions.}
The solver refines each frame until it leaves the recent window, six frames after its arrival.
The main paper reports every frame at its arrival, for comparison with existing online methods.
The last row of \Cref{tab:supp_stream_cfg} reports the same run with each frame read when it leaves the recent window.
These refined predictions are more accurate on most metrics, most clearly for video depth on Sintel and KITTI.

\input{tab/supp_stream_cfg}

\paragraph{Reuse in streaming.}
On long sequences, keeping solved frames in the window makes our model far more accurate than re-solving a short window (\cref{tab:supp_stream_abl}).
Since \dejaview diverges in the same window, this gain comes from the stationary solver rather than from the window design.
The gain grows with the sequence length, to 7-fold on ScanNet and more than 10-fold on TUM dynamics.
Keeping solved frames prevents the drift that re-solving accumulates over long sequences.
Without reuse, the solver solves each new frame together with the seven previous frames from their initial states.
Our model and \dejaview then reach nearly the same accuracy.
The L8 window keeps the solved states of its anchors, memory frames, and recent frames.
\dejaview cannot refine frames past its schedule in the same window.
Its poses diverge after about 50 frames on TUM dynamics and ScanNet.
Within the window, refreshing the frozen frames when new observations disagree with them matters most.
Without the refresh, our ATE grows more than 10-fold on TUM dynamics and ScanNet.
\input{tab/supp_stream_abl}

\paragraph{Long sequences.}
Our model streams with bounded GPU memory, since its window holds a fixed number of frames.
We track the peak GPU memory over a sequence of 2000 frames (\cref{fig:supp_memtime}).
Its GPU memory stops growing once it holds $M$ memory frames, after about 1030 frames for L8 and 260 frames for S4.
CUT3R~\cite{cut3r}, TTT3R~\cite{ttt3r}, and LingBot-Map~\cite{lingbotmap} also stay bounded.
LingBot-Map needs more than twice the GPU memory of L8.
The GPU memory of SLAM-Former~\cite{slamformer} keeps growing with the sequence.
Even on an H100 GPU with 80\,GB, offline VGGT~\cite{vggt} runs out of memory at 320 frames and StreamVGGT~\cite{streamvggt} after 365 frames.

\input{fig/supp_memtime}

\subsection{Attention across solver calls}
\label{sec:supp_attention}

\input{fig/supp_attention}

Stable predictions under continued computation do not necessarily imply that the model preserves meaningful geometric correspondences internally.
We therefore examine the global attention of the shared block across solver calls.
A head finds the correspondence of a query patch when its strongest attention in an adjacent view falls within one patch of the ground-truth correspondence.
At call 16, the same three heads are the most accurate in both models.
They find the correspondence for 51 to 74\% of the queries (\cref{fig:supp_attention}).
Our model keeps the accuracy of these heads after call 16, and their attention maps at call 64 remain close to those at call 16, with a similarity above 0.96.
The heads of \dejaview instead lose their correspondences once its schedule ends, and their accuracy drops below 3\% by call 64.

\subsection{Incremental reuse and insertion order}
\label{sec:supp_incremental}

\paragraph{Calls per insertion.}
Continuing from the current state reaches higher accuracy than re-solving from scratch at the same cost.
With 8 calls per insertion, continuing reaches an AUC@10 of 92.2 after 64 views, while re-solving with 8 calls reaches 84.8.
Re-solving needs 16 calls per insertion to reach 91.8.
Beyond 8 calls, continuing improves only slightly, to 92.9 with 16 calls and 93.1 with 32 calls.

\paragraph{Insertion order.}
We feed the same 64 views in capture order, in reverse order, and in two random orders (\cref{tab:supp_order}).
Continuing with half the calls stays within two points of re-solving in every order.

\input{tab/supp_order}

\subsection{Block-wise geometric solving}
\label{sec:supp_blocks}

The stationary update allows subsets of views to be refined without synchronizing the computational progress of the entire reconstruction.
As shown in \cref{fig:supp_blocks}, successive block-wise updates progressively refine the scene toward the geometry obtained by joint solving, while the remaining views provide fixed geometric context.
For 128 views, block-wise solving peaks at 1.38\,GiB of GPU memory, compared with 4.39\,GiB for joint solving ($3\times$) and 13.98\,GiB for VGGT ($10\times$).

\input{fig/supp_blocks}

\subsection{Bridging views as new constraints}
\label{sec:supp_bridge}

Additional solver calls do not introduce new geometric constraints, and may therefore leave an ambiguous reconstruction unresolved.
\Cref{fig:supp_bridge_qual} illustrates this distinction.
Continued refinement alone leaves the two weakly overlapping view groups misaligned, whereas bridging views constrain their relative pose and allow our model to correct the existing reconstruction.
In contrast, \dejaview fails to correct the same reconstruction through continuation, despite receiving the same additional views.

\input{fig/supp_bridge_qual}

\subsection{Failure case}
\label{sec:supp_limitations}
Although our solver can continue refining a reconstruction, additional solver calls alone do not guarantee accurate global geometry in underconstrained settings.
We illustrate this with a window from the stopping analysis (\cref{sec:supp_convergence,fig:supp_fail}), the only one among the 144 stopped windows whose rotation error exceeds $5^\circ$.
In this window, two groups of views have little overlap, making reconstruction challenging for the compared GFMs as well.
Further solver calls do not correct the relative pose of the two groups, as the available constraints remain unchanged.
However, as demonstrated in \cref{sec:supp_bridge}, new observations providing the missing constraints can enable the solver to correct the reconstruction through continued solving.

\input{fig/supp_fail}

\section{Additional results}
\label{sec:supp_results}

We complement the main benchmarks with an evaluation on large aerial
scenes and qualitative results under different view-arrival patterns,
including offline and streaming reconstruction.

\subsection{Large aerial scenes}
\label{sec:supp_quantitative}

Our model estimates the most accurate cameras on both large aerial scenes of Mill19~\cite{meganerf}, Rubble and Building (\cref{tab:supp_mill19}).
Every method reconstructs each scene offline from the same 96 evenly spaced images in a single window.
We compare the predicted cameras with the reference poses of the dataset.

\input{tab/supp_mill19}

\subsection{Qualitative results}
\label{sec:supp_qualitative}

\paragraph{Arrival patterns.}
The same solver reconstructs a scene whether its views arrive all at once, in batches, one at a time, or in arbitrary order (\cref{fig:supp_join_stages}).
Views that join continue the running solve instead of restarting it.
All four patterns end in nearly the same accurate reconstruction of a scene (\cref{fig:supp_join_same}).

\input{fig/supp_join_stages}

\input{fig/supp_join_same}

\paragraph{Offline reconstruction.}
\Cref{fig:supp_qual_scannetpp,fig:supp_qual_eth} present two large indoor scenes, where every method~\cite{burzio2026dejaview,vggt,pi3,depthanything3,vggtomega} sees all images at once.

\paragraph{Streaming reconstruction.}
\Cref{fig:supp_stream_qual} presents four more sequences streamed with the default configuration (L8), each frame shown with its prediction at arrival.

\input{fig/supp_qual_scannetpp}

\input{fig/supp_qual_eth}

\input{fig/supp_stream_qual}

%% file: tab/supp_training_data.tex
\begin{table}[t]
\begin{center}
\resizebox{\linewidth}{!}{%
\fontsize{7.5}{9}\selectfont
\begin{tabular*}{\linewidth}{@{\extracolsep{\fill}}llr@{}}
\toprule
Type & Dataset & Weight (\%) \\
\midrule
\multirow{4}{*}{Indoor} & ScanNet++~\cite{scannetpp} & 10.8 \\
 & ScanNet~\cite{scannet} & 7.4 \\
 & CA-1M~\cite{ca1m} & 3.3 \\
 & Taskonomy~\cite{taskonomy} & 3.3 \\
\midrule
\multirow{5}{*}{Object} & CO3D~\cite{co3d} & 7.4 \\
 & NAVI~\cite{navi} & 3.3 \\
 & Skoltech3D~\cite{skoltech3d} & 2.5 \\
 & OmniObject3D~\cite{omniobject3d} & 2.5 \\
 & Objaverse renders~\cite{objaverse} & 2.5 \\
\midrule
\multirow{5}{*}{Outdoor} & DL3DV~\cite{dl3dv} & 8.2 \\
 & BlendedMVS~\cite{blendedmvs} & 4.1 \\
 & MegaDepth~\cite{megadepth} & 3.3 \\
 & Oxford Spires~\cite{oxfordspires} & 1.6 \\
 & Mip-NeRF 360$^\ast$~\cite{mipnerf360} & 0.8 \\
\midrule
\multirow{4}{*}{Driving} & Waymo$^\ast$~\cite{waymo} & 4.1 \\
 & Argoverse 2~\cite{argoverse2} & 3.3 \\
 & ParallelDomain-4D~\cite{pd4d} & 3.3 \\
 & Virtual KITTI 2~\cite{vkitti2} & 1.6 \\
\midrule
\multirow{7}{*}{Synthetic} & Infinigen~\cite{infinigen} & 9.1 \\
 & TartanAir~\cite{tartanair} & 5.8 \\
 & Hypersim~\cite{hypersim} & 5.8 \\
 & MVS-Synth~\cite{mvssynth} & 1.6 \\
 & Spring~\cite{spring} & 1.6 \\
 & MatrixCity~\cite{matrixcity} & 1.6 \\
 & UnrealStereo4K~\cite{unrealstereo4k} & 1.2 \\
\bottomrule
\end{tabular*}%
}
\end{center}
\caption{\textbf{Training data.}
Sampling weights (\%) in the second training stage.
$^\ast$Supervised only by the distillation loss.}
\label{tab:\currfilebase}
\end{table}

%% file: fig/supp_stopdist.tex
\begin{figure}[t]
\centering
\includegraphics[width=\linewidth]{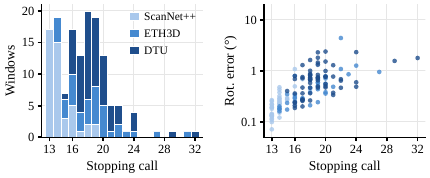}
\caption{\textbf{Stopping calls.}
Left: number of windows per stopping call, for the 144 of the 148 windows of the stability analysis that reach $\gamma=0.02$, stacked by dataset.
Right: rotation error of each window at its stopping call.}
\label{fig:supp_stopdist}
\end{figure}

%% file: tab/supp_tau.tex
\begin{table}[t]
\begin{center}
\resizebox{\linewidth}{!}{%
\setlength{\tabcolsep}{3pt}%
\fontsize{7.5}{9}\selectfont
\begin{tabular}{@{}lcccccccccc@{}}
\toprule
& \multicolumn{2}{c}{DTU~\cite{dtu}} & \multicolumn{2}{c}{ETH3D~\cite{eth3d}} & \multicolumn{2}{c}{7-Scenes~\cite{7scenes}} & \multicolumn{2}{c}{ScanNet++~\cite{scannetpp}} & \multicolumn{2}{c}{nuScenes~\cite{nuscenes}} \\
\cmidrule(lr){2-3}\cmidrule(lr){4-5}\cmidrule(lr){6-7}\cmidrule(lr){8-9}\cmidrule(lr){10-11}
$\gamma$ & AUC & Calls & AUC & Calls & AUC & Calls & AUC & Calls & AUC & Calls \\
\midrule
0.01 & 85.9 & 36.1 & 69.8 & 54.9 & 14.2 & 35.8 & 81.5 & 39.8 & 41.8 & 31.1 \\
0.02 & 85.5 & 19.7 & 70.4 & 20.5 & 14.2 & 22.6 & 81.2 & 17.3 & 41.5 & 18.2 \\
0.05 & 81.5 & 13.0 & 62.0 & 13.5 & 13.9 & 12.9 & 77.0 & 11.8 & 40.9 & 11.9 \\
\bottomrule
\end{tabular}%
}
\end{center}
\caption{\textbf{Sensitivity to the stopping threshold $\gamma$.}
Pose AUC@3 and the mean number of solver calls per scene.}
\label{tab:\currfilebase}
\end{table}

%% file: tab/supp_hard.tex
\begin{table}[t]
\begin{center}
\resizebox{\linewidth}{!}{%
\setlength{\tabcolsep}{2pt}%
\fontsize{7.5}{9}\selectfont
\begin{tabular}{@{}lccccccccc@{}}
\toprule
& & & & \multicolumn{2}{c}{Error ($^\circ$)} & Stops & \multicolumn{2}{c}{Start diff.\ ($^\circ$)} & \dejaview \\
\cmidrule(lr){5-6}\cmidrule(lr){8-9}
Windows & \# & Stopped & Calls & Stop & 128 & $>5^\circ$ & Noise & Swap & at 128 ($^\circ$) \\
\midrule
Stability analysis & 148 & \phantom{0}97\% & 17.5 & 0.44 & 0.42 & 1 & 0.09 & 0.11 & 96.0 \\
\midrule
Tanks and Temples & \phantom{0}39 & 100\% & 17.7 & 0.29 & 0.27 & 0 & 0.07 & 0.09 & 86.6 \\
7-Scenes & \phantom{0}36 & \phantom{0}94\% & 16.2 & 1.56 & 1.57 & 0 & 0.06 & 0.06 & 82.9 \\
TUM dynamics & \phantom{0}32 & 100\% & 15.7 & 0.42 & 0.39 & 0 & 0.04 & 0.05 & 75.1 \\
\bottomrule
\end{tabular}%
}
\end{center}
\caption{\textbf{Stopping on other datasets.}
Eight-view windows solved for 128 calls.
Stopped is the share of windows that reach $\gamma=0.02$, and Calls is their mean stopping call.
Error is the median rotation error of the stopped windows at the stopping call and at call 128.
Stops $>5^\circ$ counts the stopped windows with an error above $5^\circ$ at the stopping call.
Start diff.\ is the median rotation difference after 128 calls between the start from $\mathbf{z}^0$ and the noisy or swapped start.
The last column is the median rotation error of \dejaview at call 128.}
\label{tab:\currfilebase}
\end{table}

%% file: fig/supp_warminit.tex
\begin{figure}[t]
\centering
\includegraphics[width=\linewidth]{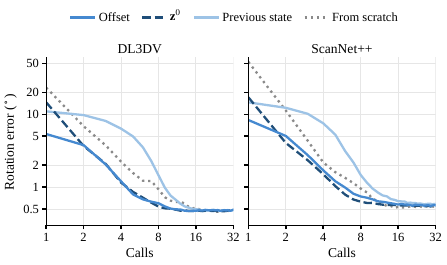}
\caption{\textbf{Initialization of new views.}
Median rotation error of a newly arrived frame over solver calls, for 656 frames per dataset, 41 from each of 16 sequences.
The new frame starts from its initial state $\mathbf{z}^0$, from $\mathbf{z}^0$ plus the solved offset of the previous frame (Offset), or from the solved state of the previous frame (Previous state).
The frames that remain in the window keep their solved states.
``From scratch'' solves the whole window from the initial states.}
\label{fig:\currfilebase}
\end{figure}

%% file: fig/supp_streamwarm.tex
\begin{figure}[t]
\centering
\includegraphics[width=\linewidth]{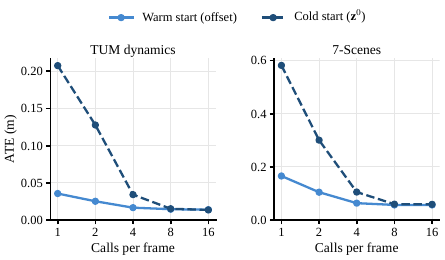}
\caption{\textbf{Initialization in streaming.}
ATE of on-arrival predictions with the L8 window for different numbers of calls per frame.
Each new frame starts from its initial state $\mathbf{z}^0$ (cold start) or from $\mathbf{z}^0$ plus the solved offset of the previous frame (warm start).}
\label{fig:\currfilebase}
\end{figure}

%% file: tab/supp_stream_cfg.tex
\begin{table*}[t]
\begin{center}
\resizebox{\textwidth}{!}{%
\setlength{\tabcolsep}{3pt}%
\fontsize{7.5}{9}\selectfont
\begin{tabular*}{\linewidth}{@{\extracolsep{\fill}}lcccccccccccccc@{}}
\toprule
& & & & & \multicolumn{3}{c}{ATE (m)$\downarrow$} & \multicolumn{3}{c}{Video depth AbsRel$\downarrow$} & \multicolumn{2}{c}{Chamfer (m)$\downarrow$} & Time$\downarrow$ & GPU mem.$\downarrow$ \\
\cmidrule(lr){6-8}\cmidrule(lr){9-11}\cmidrule(lr){12-13}
Config & $a$ & $r$ & $M$ & $m$ & Sintel & TUM & ScanNet & Sintel & Bonn & KITTI & 7-Scenes & NRGBD & (ms) & (GB) \\
\midrule
L8 & 8 & 6 & 64 & 8 & .157 & .015 & .077 & .322 & .064 & .078 & .018 & .018 & 95 & 5.0 \\
L4 & 8 & 6 & 64 & 4 & .163 & .017 & .091 & .321 & .064 & .083 & .037 & .024 & 65 & 5.0 \\
S4 & 4 & 4 & 16 & 4 & .179 & .017 & .180 & .328 & .065 & .088 & .072 & .045 & 51 & 2.1 \\
\midrule
L8, refined & 8 & 6 & 64 & 8 & .162 & .014 & .076 & .302 & .065 & .072 & .017 & .017 & 95 & 5.0 \\
\bottomrule
\end{tabular*}%
}
\end{center}
\caption{\textbf{Streaming configurations.}
$a$, $r$, $M$, and $m$ are the numbers of anchors, recent frames, memory frames, and calls per frame.
Every configuration takes a memory frame every 16 frames.
Runtime is the mean time per frame on an H100 GPU, including the image encoder and the prediction heads.
As in the main paper, we measure it on the TUM dynamics sequences sitting\_halfsphere, sitting\_static, and walking\_rpy.
The last row reads each frame of the L8 run when it leaves the recent window instead of at its arrival.}
\label{tab:\currfilebase}
\end{table*}

%% file: tab/supp_stream_abl.tex
\begin{table}[t]
\begin{center}
\resizebox{\linewidth}{!}{%
\setlength{\tabcolsep}{5pt}%
\fontsize{7.5}{9}\selectfont
\begin{tabular}{@{}llcc@{}}
\toprule
Window & Method & TUM (878) & ScanNet (602) \\
\midrule
No reuse & \dejaview & 0.218 & 0.560 \\
& Solv3R & 0.214 & 0.574 \\
\midrule
L8 & \dejaview & div. & div. \\
& Solv3R & 0.015 & 0.077 \\
& \quad no refresh & 0.160 & 1.05\phantom{0} \\
& \quad no recent window & 0.014 & 0.324 \\
\bottomrule
\end{tabular}%
}
\end{center}
\caption{\textbf{Reuse in streaming.}
ATE (m) of on-arrival predictions on the long sequences of TUM dynamics and ScanNet.
Without reuse, the solver solves each new frame with the seven previous frames from their initial states for 16 calls.
We chain its pose through the first frame of the window.
In the L8 window, \dejaview runs its step schedule for each frame.
Its poses diverge (div.).
Without the refresh, the frozen frames never rejoin the update.
No recent window drops the six recent frames, so that each call updates only the new frame.}
\label{tab:\currfilebase}
\end{table}

%% file: fig/supp_memtime.tex
\begin{figure}[t]
\centering
\includegraphics[width=\linewidth]{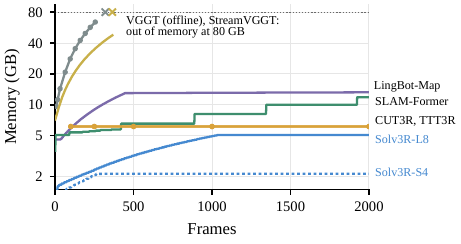}
\caption{\textbf{Memory on a long sequence.}
GPU memory over the 2000 frames of the long office sequence of TUM RGB-D, on a logarithmic axis.
All methods are evaluated on a single H100 GPU.
Crosses mark where a method runs out of the 80\,GB of GPU memory (gray dotted line).}
\label{fig:supp_memtime}
\end{figure}

%% file: fig/supp_attention.tex
\begin{figure}[t]
\centering
\includegraphics[width=\linewidth]{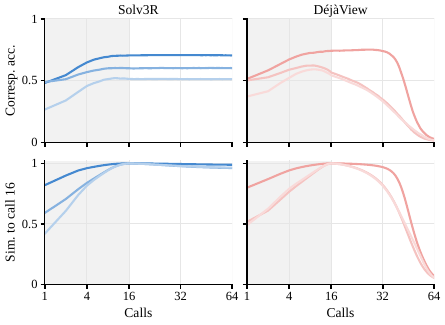}
\caption{\textbf{Attention across solver calls.}
We show the three global-attention heads, out of 12, that best find ground-truth correspondences at call 16.
These are the same heads in our model (left) and \dejaview (right).
Top: fraction of query patches whose strongest attention in an adjacent view falls within one patch of the ground-truth correspondence.
Bottom: similarity to the call-16 attention map, measured as the share of the call-16 attention mass on the 64 most attended keys of each call and normalized to 1 at call 16.
The shaded range is the 16-step schedule of \dejaview.
Averaged over 82 windows of ETH3D and ScanNet++.}
\label{fig:\currfilebase}
\end{figure}

%% file: tab/supp_order.tex
\begin{table}[t]
\begin{center}
\resizebox{\linewidth}{!}{%
\setlength{\tabcolsep}{4pt}%
\fontsize{7.5}{9}\selectfont
\begin{tabular*}{\linewidth}{@{\extracolsep{\fill}}lcccc@{}}
\toprule
& \multicolumn{2}{c}{DL3DV~\cite{dl3dv}} & \multicolumn{2}{c}{ScanNet++~\cite{scannetpp}} \\
\cmidrule(lr){2-3}\cmidrule(lr){4-5}
Order & Continue & Re-solve & Continue & Re-solve \\
\midrule
Capture & 92.0 & 91.2 & 92.4 & 92.5 \\
Reverse & 89.9 & 91.6 & 92.4 & 92.4 \\
Random 1 & 91.7 & 91.7 & 92.1 & 91.8 \\
Random 2 & 92.0 & 92.1 & 92.3 & 92.3 \\
\bottomrule
\end{tabular*}%
}
\end{center}
\caption{\textbf{Sensitivity to the insertion order.}
AUC@10 after inserting 64 views, 8 at a time.
Continue runs 8 calls per insertion from the current state, and re-solve runs 16 calls from scratch.}
\label{tab:\currfilebase}
\end{table}

%% file: fig/supp_blocks.tex
\begin{figure*}[t]
\centering
\includegraphics[width=\linewidth]{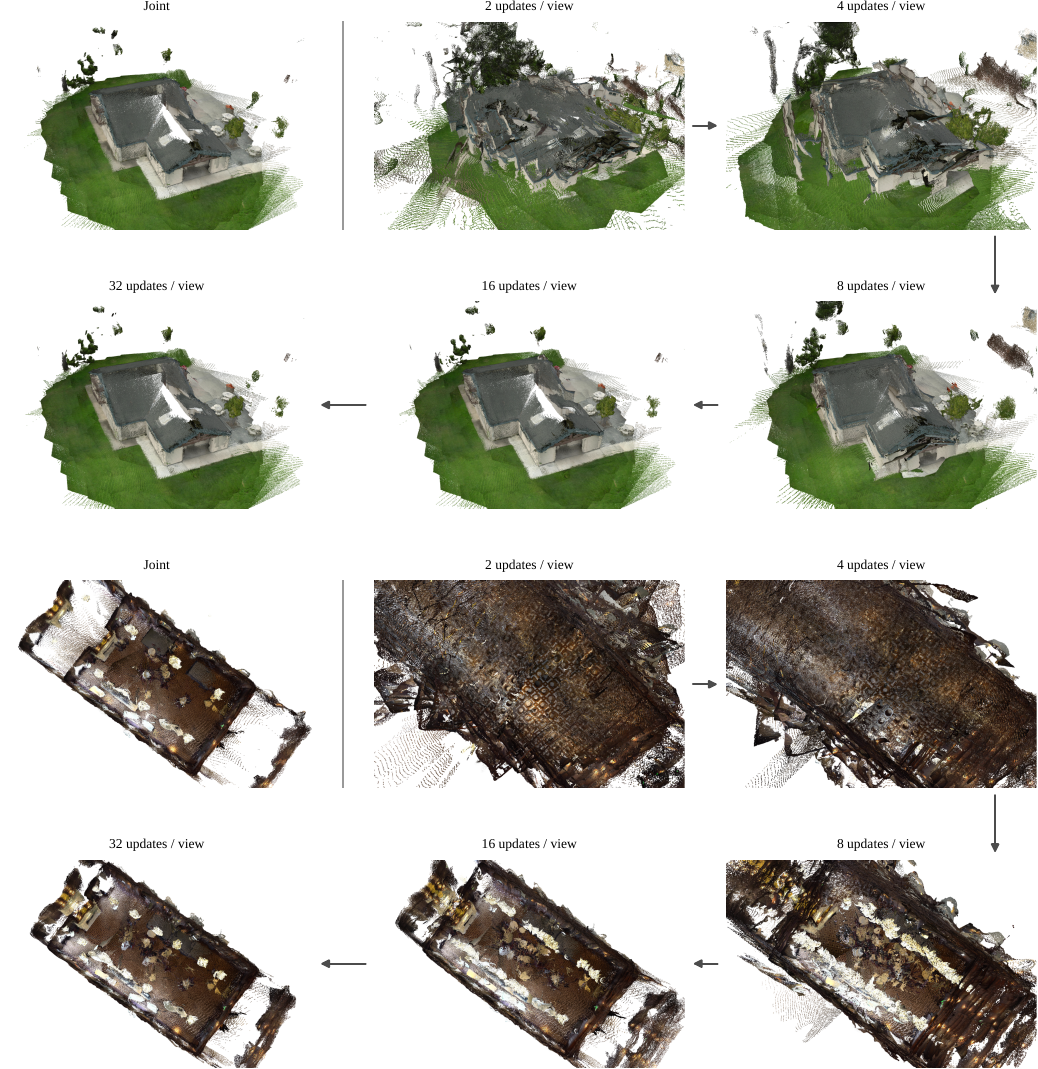}
\caption{\textbf{Block-wise solving.}
Our reconstructions of two scenes~\cite{tanksandtemples} after 2 to 32 updates per view on average with random blocks of eight views, in the order of the arrows, next to the joint solve.
All panels of a scene share the same viewpoint.
Block-wise updates progressively refine the reconstruction toward the joint solve, although each call updates only a subset of views.}
\label{fig:\currfilebase}
\end{figure*}

%% file: fig/supp_bridge_qual.tex
\begin{figure*}[t]
\centering
\includegraphics[width=\linewidth]{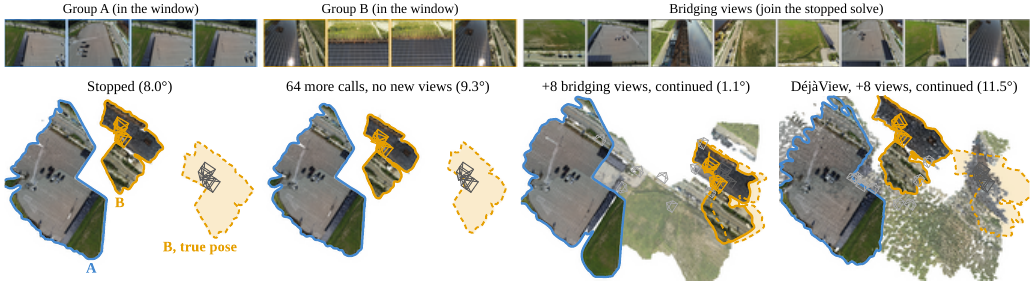}
\caption{\textbf{Bridging views as new constraints.}
The two view groups of this aerial window~\cite{meganerf} have almost no co-visibility, so the images provide insufficient constraints to reliably recover their relative pose, and the solve stops with group B misplaced.
Further calls add no constraints and leave it misplaced.
Bridging views that observe both groups add the missing constraints, and continuing from the stopped state places group B correctly.
\dejaview, given the same views, does not correct it.
Seen from above, with group A in the ground-truth frame and the true pose of group B as a dashed outline.}
\label{fig:\currfilebase}
\end{figure*}

%% file: fig/supp_fail.tex
\begin{figure}[t]
\centering
\includegraphics[width=0.9\linewidth]{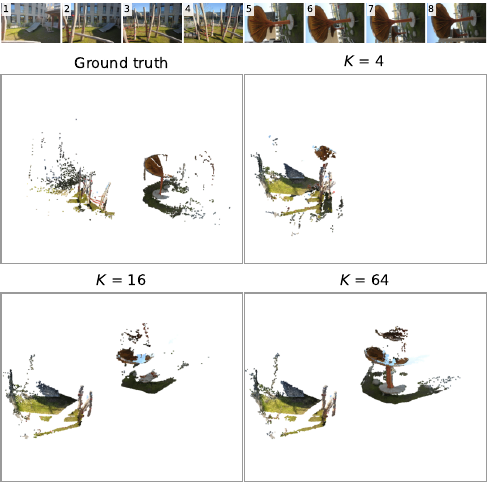}
\caption{\textbf{Failure case.}
In this window, views 5--8 have little overlap with views 1--4.
Our model places them at a wrong position and orientation within 16 calls and keeps them there up to call 64.}
\label{fig:supp_fail}
\end{figure}

%% file: tab/supp_mill19.tex
\begin{table}[t]
\begin{center}
\resizebox{\linewidth}{!}{%
\setlength{\tabcolsep}{4pt}%
\fontsize{7.5}{9}\selectfont
\begin{tabular*}{\linewidth}{@{\extracolsep{\fill}}lcccc@{}}
\toprule
& \multicolumn{2}{c}{Rubble} & \multicolumn{2}{c}{Building} \\
\cmidrule(lr){2-3}\cmidrule(lr){4-5}
Method & @3 & @30 & @3 & @30 \\
\midrule
VGGT~\cite{vggt} & \phantom{0}5.3 & 82.4 & \phantom{0}2.5 & 70.6 \\
VGGT-$\Omega$~\cite{vggtomega} & 34.6 & 91.8 & 14.6 & 86.4 \\
$\pi^3$~\cite{pi3} & 14.9 & 88.3 & \phantom{0}1.8 & 75.9 \\
DA3-L~\cite{depthanything3} & \phantom{0}3.8 & 74.0 & \phantom{0}1.1 & 55.8 \\
\dejaview~\cite{burzio2026dejaview} & 39.4 & 93.0 & 11.0 & 87.0 \\
\midrule
\textbf{Solv3R} & \textbf{48.9} & \textbf{94.1} & \textbf{22.5} & \textbf{89.6} \\
\bottomrule
\end{tabular*}%
}
\end{center}
\caption{\textbf{Large aerial scenes (Mill19).}
Camera pose AUC@3 and AUC@30 from 96 images per scene.}
\label{tab:\currfilebase}
\end{table}

%% file: fig/supp_join_stages.tex
\begin{figure*}[p]
\centering
\includegraphics[width=\linewidth]{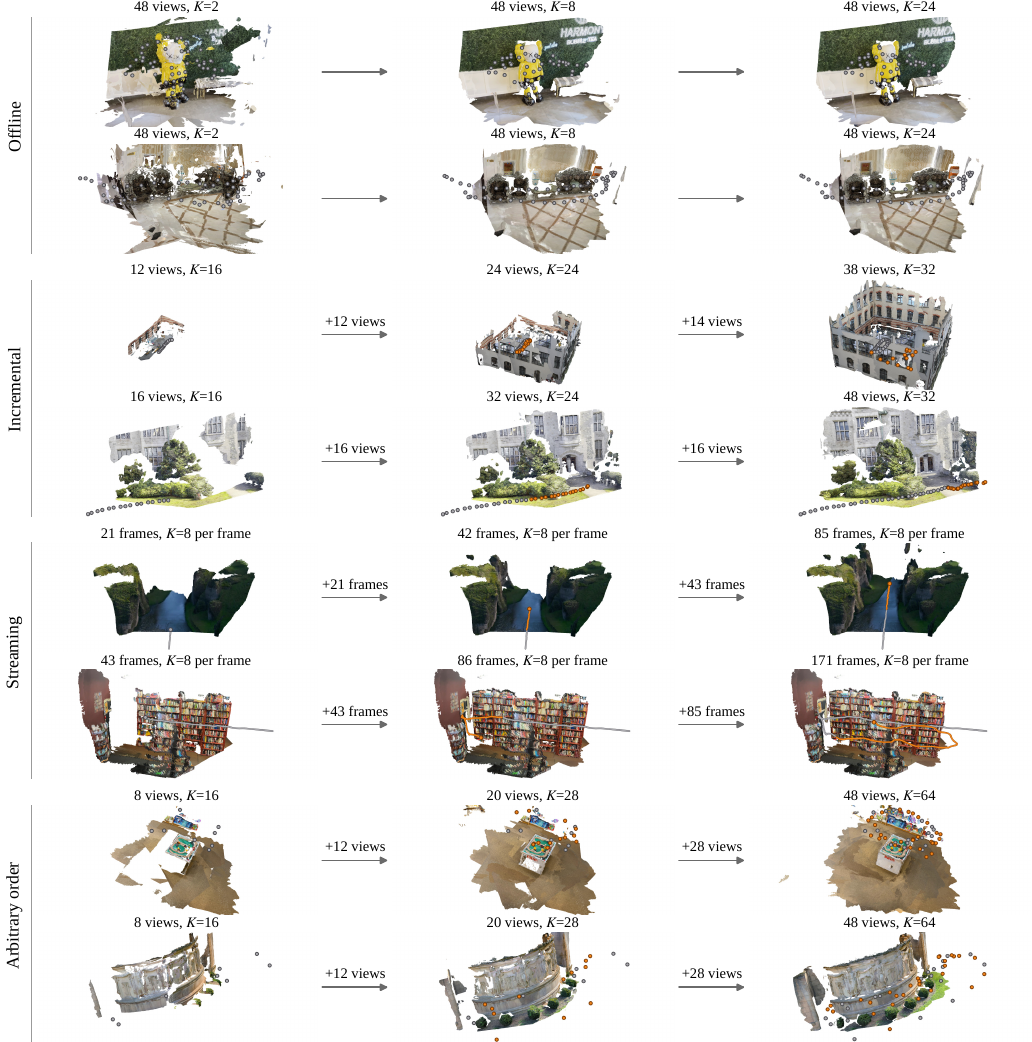}
\caption{\textbf{Reconstructions that grow as views arrive.}
Each row shows one scene at three stages of one arrival pattern.
Offline solves all views jointly, incremental adds the capture in thirds, streaming adds video frames one at a time, and arbitrary order adds random groups of views.
Views that join continue the running solve, and the views already in it keep their states.
Above each panel are the number of views in the solve and the number $K$ of solver calls run so far.
Arrows give the views that joined, and orange marks their cameras.}
\label{fig:supp_join_stages}
\end{figure*}

%% file: fig/supp_join_same.tex
\begin{figure*}[t]
\centering
\includegraphics[width=\linewidth]{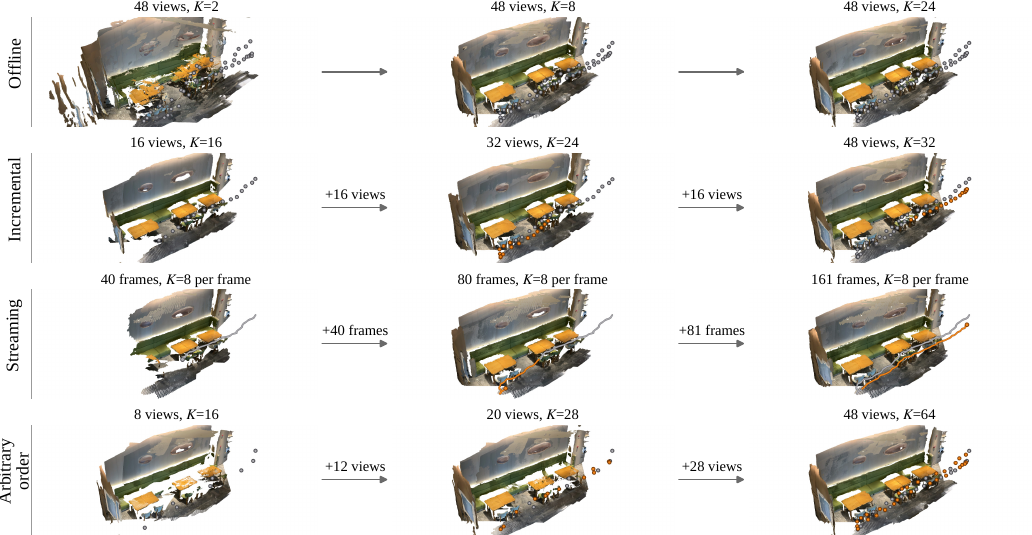}
\caption{\textbf{One scene under the four arrival patterns.}
Shown as in \cref{fig:supp_join_stages}.
All four patterns end in nearly the same reconstruction, although the views arrive in different orders and receive different numbers of calls.}
\label{fig:supp_join_same}
\end{figure*}

%% file: fig/supp_qual_scannetpp.tex
\begin{figure*}[t]
\centering
\includegraphics[width=\linewidth]{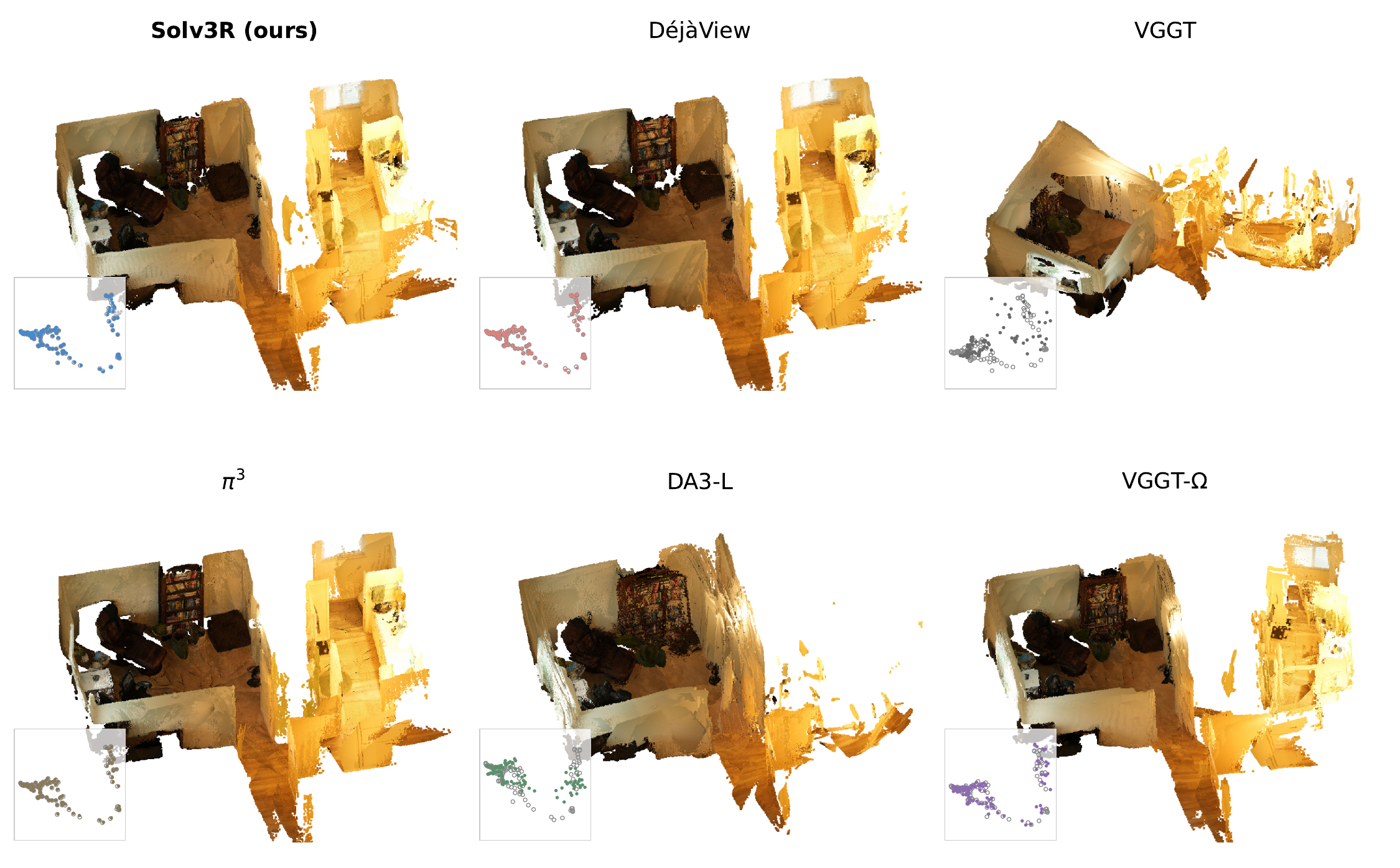}
\caption{\textbf{Offline reconstruction comparison.}
All methods see the same 100 images at once.}
\label{fig:supp_qual_scannetpp}
\end{figure*}

%% file: fig/supp_qual_eth.tex
\begin{figure*}[t]
\centering
\includegraphics[width=0.9\linewidth]{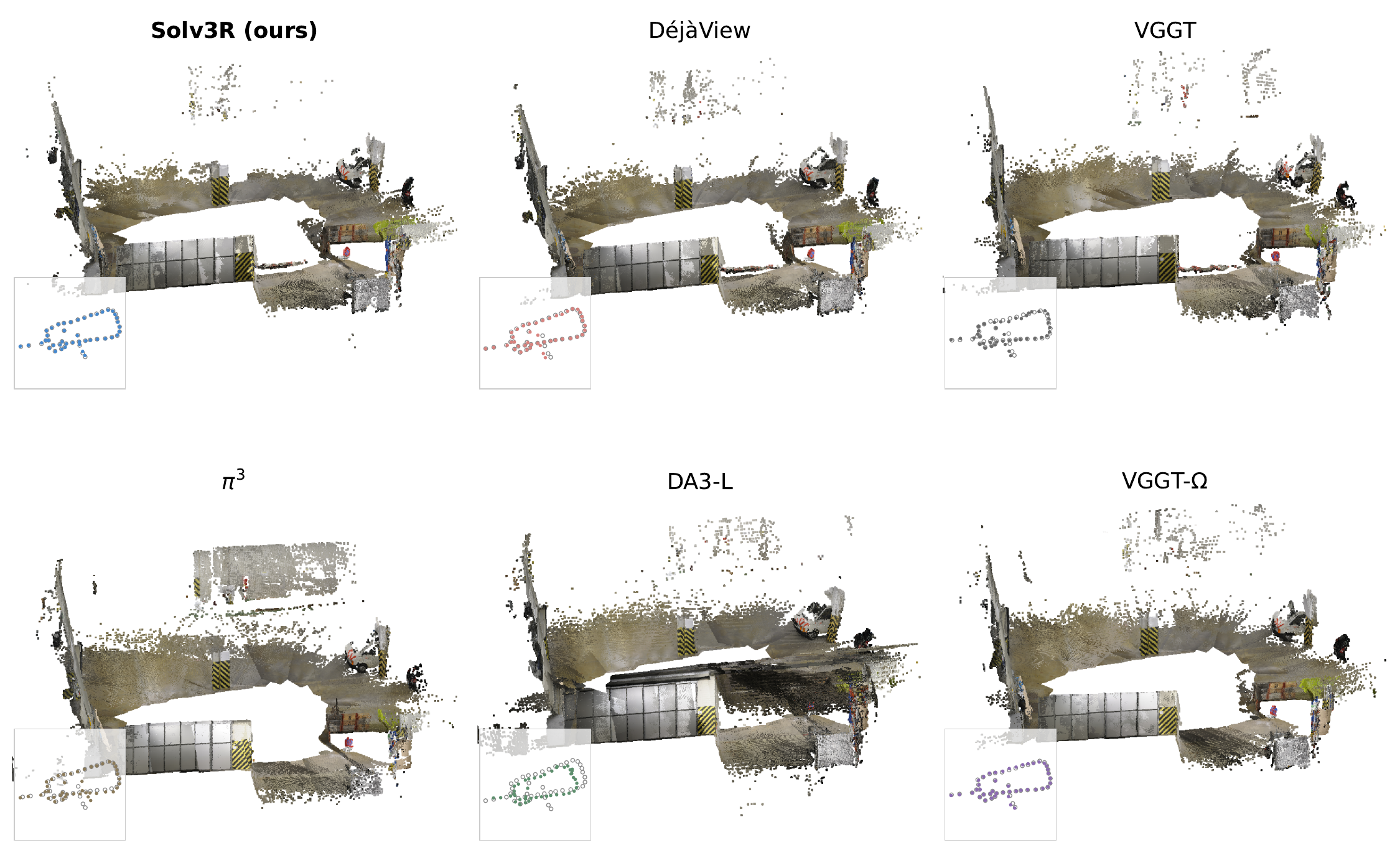}
\caption{\textbf{Offline reconstruction comparison.}
All methods see the same 44 images at once.}
\label{fig:supp_qual_eth}
\end{figure*}

%% file: fig/supp_stream_qual.tex
\begin{figure*}[t]
\centering
\includegraphics[width=\linewidth]{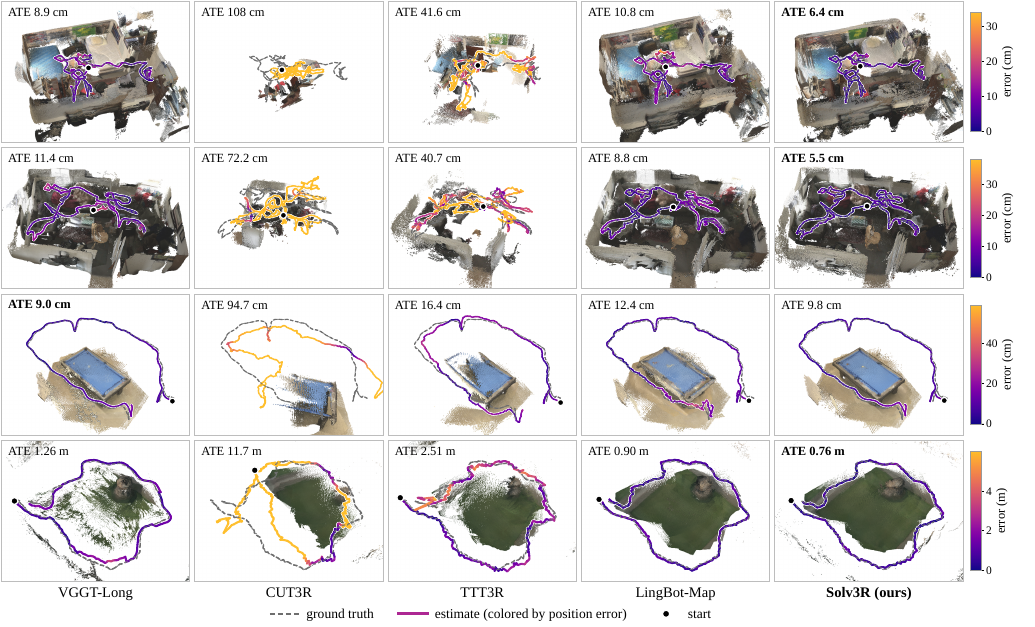}
\caption{\textbf{Streaming reconstruction comparison.}
The methods and settings are those of the streaming figure of the main paper.
Camera trajectories are colored by position error.}
\label{fig:supp_stream_qual}
\end{figure*}